\documentclass[preprint,3p,times]{elsarticle}
\RequirePackage{multirow,booktabs,subfigure,color,array,hhline,makecell}
\usepackage{amssymb}
\usepackage{amsmath}
\usepackage{graphicx}
\usepackage{amsthm}
\usepackage{mathrsfs}
\usepackage{indentfirst}
\usepackage[colorlinks,citecolor=blue,urlcolor=blue]{hyperref}
\allowdisplaybreaks
\usepackage[table]{xcolor}
\usepackage{tikz-network}
\usepackage{pgf}
\usepackage{tikz}
\usepackage{subfigure}
\usetikzlibrary{arrows, decorations.pathmorphing, backgrounds, positioning, fit, petri, automata}
\usetikzlibrary{shadows,arrows,positioning}
\RequirePackage{algorithm,algpseudocode}
\allowdisplaybreaks
\usepackage{changes}

\newtheorem{thm}{Theorem}

\newtheorem{lem}{Lemma}
\newtheorem{assum}{Assumption}
\newtheorem{rem}{Remark}
\newtheorem{cor}{Corollary}

\allowdisplaybreaks[4]
\AtBeginDocument{%
	\providecommand\BibTeX{{%
			\normalfont B\kern-0.5em{\scshape i\kern-0.25em b}\kern-0.8em\TeX}}}
\journal{~}
\begin{document}
\begin{frontmatter}
\title{Latent class analysis for multi-layer categorical data}

\author[label1]{Huan Qing\corref{cor1}}
\ead{qinghuan@cqut.edu.cn \&qinghuan@u.nus.edu \&qinghuan07131995@163.com}
\cortext[cor1]{Corresponding author.}
\address[label1]{School of Economics and Finance, Lab of Financial Risk Intelligent Early Warning and Modern Governance,  Chongqing University of Technology, Chongqing, 400054, China}
\begin{abstract}
Traditional categorical data, often collected in psychological tests and educational assessments, are typically single-layer and gathered only once. This paper considers a more general case, multi-layer categorical data with polytomous responses. To model such data, we present a novel statistical model, the multi-layer latent class model (multi-layer LCM). This model assumes that all layers share common subjects and items. To discover subjects' latent classes and other model parameters under this model, we develop three efficient spectral methods based on the sum of response matrices, the sum of Gram matrices, and the debiased sum of Gram matrices, respectively. Within the framework of multi-layer LCM, we demonstrate the estimation consistency of these methods under mild conditions regarding data sparsity. Our theoretical findings reveal two key insights: (1) increasing the number of layers can enhance the performance of the proposed methods, highlighting the advantages of considering multiple layers in latent class analysis; (2) we theoretically show that the algorithm based on the debiased sum of Gram matrices usually performs best. Additionally, we propose an approach that combines the averaged modularity metric with our methods to determine the number of latent classes. Extensive experiments are conducted to support our theoretical results and show the ‌powerfulness of our methods in the task of learning latent classes and estimating the number of latent classes in multi-layer categorical data with polytomous responses.
\end{abstract}
\end{frontmatter}
\section{Introduction}\label{sec1}
Categorical data, which records subjects' responses to various items, is widely collected in psychological tests and educational assessments \citep{sloane1996introduction,agresti2012categorical,chen2019joint,shang2021partial}. When responses are limited to two options, they are called binary, often representing yes/no or correct/wrong answers. On the other hand, responses with more than two outcomes are called polytomous, including multiple choices such as a/b/c/d/e. Traditionally, categorical data is gathered only once and represented mathematically by an $N\times J$ matrix, denoted as $R$, where $N$ represents the number of subjects (participants or individuals), $J$ represents the number of items (questions), and $R(i,j)$ designates the observed response of subject $i$ to the $j$-th item. The elements of matrix $R$ range from $\{0,1,\ldots, M\}$ with $M$ representing the maximum number of possible responses. In binary scenarios, $M$ is equal to 1, while in polytomous cases, $M$ is greater than or equal to 2, with 0 indicating no response and values from $\{1,2,\ldots, M\}$ representing different outcomes.

\begin{figure}
\centering
\resizebox{\columnwidth}{!}{
\subfigure[]{\includegraphics[width=0.2\textwidth]{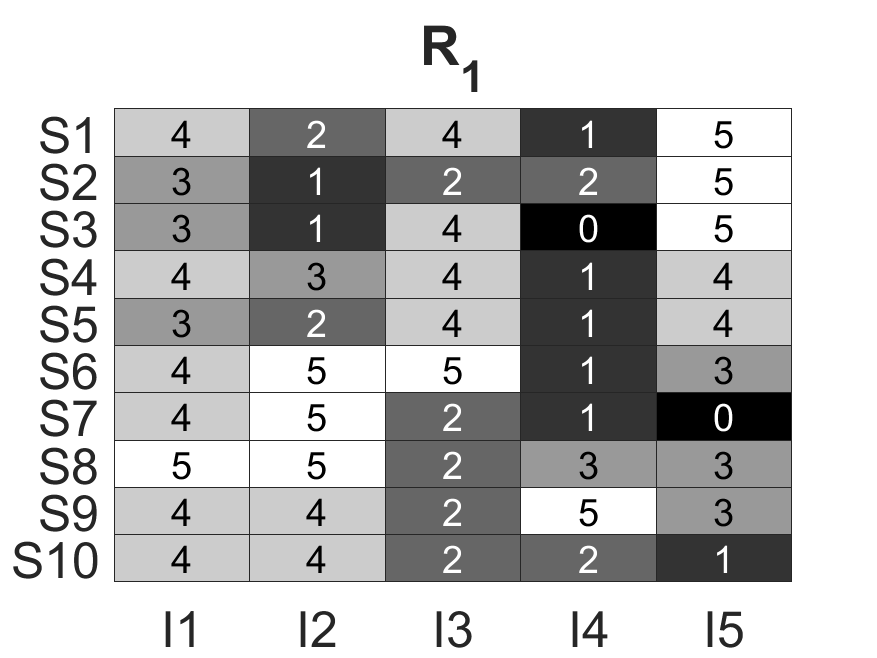}}
\subfigure[]{\includegraphics[width=0.2\textwidth]{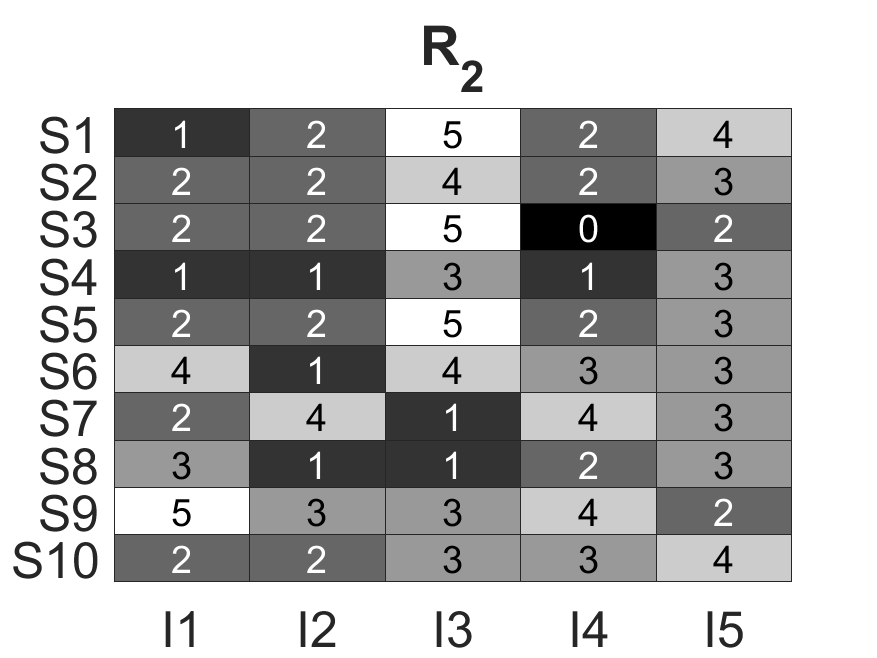}}
\subfigure[]{\includegraphics[width=0.2\textwidth]{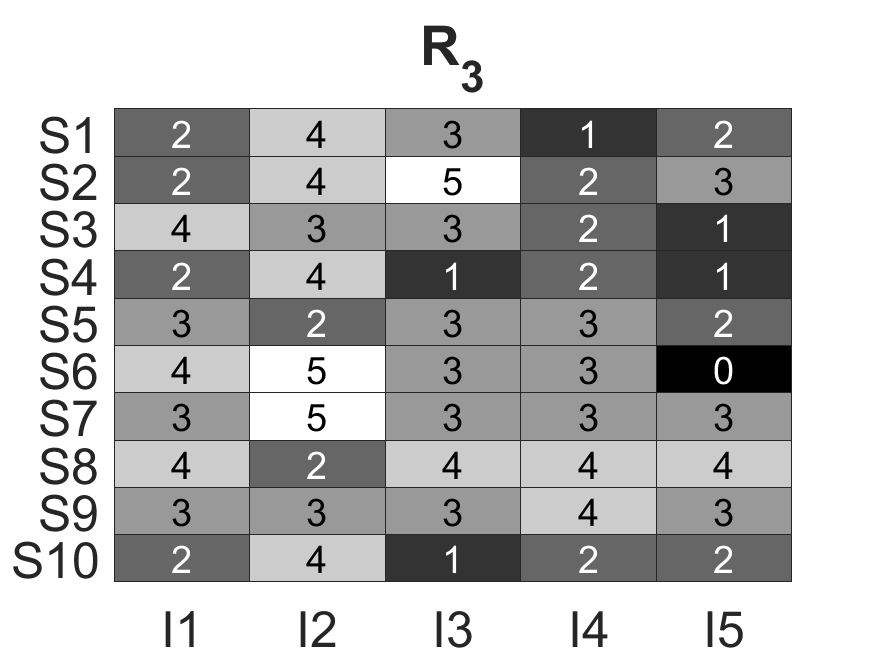}}
}
\caption{A simple example of multi-layer categorical data with 10 subjects, 5 items, and 3 layers. Here, S$i$ denotes subject $i$ for $i\in\{1,2,\ldots,10\}$, I$j$ denotes item $j$ for $j\in\{1,2,\ldots,5\}$, and $R_{l}\in\{0,1,2,\ldots,5\}^{10\times 5}$ denotes the $l$-th response matrix for $l\in\{1,2,3\}$.}
\label{R6} %% label for entire figure
\end{figure}

In this paper, we consider a more general case where the tests/assessments/surveys are conducted multiple times, resulting in multi-layer categorical data with polytomous responses. For instance, a group of high school students engages in an annual psychological questionnaire survey, resulting in multi-layer categorical data over time. Likewise, in educational evaluation, students may participate in the same educational survey questionnaire periodically, facilitating longitudinal analysis. Similarly, voters also have the opportunity to participate in the same political questionnaire survey at various time intervals, allowing for a comprehensive understanding of their evolving perspectives. It is easy to see that such data has common subjects (individuals) and common items (questions) shared by all layers. Figure \ref{R6} presents a simple example of multi-layer categorical data with polytomous responses, highlighting the varying responses of subjects to identical items across different layers. When the data only has one layer, we call it single-layer categorical data.

The latent class model (LCM) \citep{goodman1974exploratory} is a powerful tool for describing single-layer categorical data with hidden subgroup structures. This model presupposes that subjects in the same subgroups have similar response patterns. Multiple methodologies have been designed to discover the latent classes of subjects within the LCM framework, including Bayesian inference techniques \citep{garrett2000latent,asparouhov2011using,white2014bayeslca,li2018bayesian}, maximum likelihood estimation methods \citep{bakk2016robustness,chen2022beyond}, a tensor-based approach \citep{zeng2023tensor}, and spectral clustering methods \citep{qing2023latent}. Nevertheless, these approaches are tailored for single-layer categorical data and are inapplicable to identify latent classes within multi-layer categorical datasets. Because multi-layer categorical data encompasses more responses than single-layer categorical data, it should potentially facilitate a deeper understanding of the underlying latent classes among subjects.

We study the problem of latent class analysis in multi-layer categorical data with polytomous responses. This problem is closely related to the community detection task in multi-layer networks, where the spectral clustering methods with theoretical guarantees are adopted in multi-layer undirected networks \citep{han2015consistent,paul2020spectral,lei2023bias} and multi-layer directed networks \citep{su2024spectral}. For multi-layer undirected networks, spectral methods using the sum of adjacency matrices are studied in \citep{han2015consistent,paul2020spectral}, while a novel debiased spectral clustering approach is introduced in \citep{lei2023bias}. The work presented in \citep{su2024spectral} extends the methodology of \citep{lei2023bias} to multi-layer directed networks. However, adapting spectral methods from multi-layer networks to multi-layer categorical data with polytomous responses remains an unresolved challenge. In this paper, our main contributions are summarized as follows:
\begin{itemize}
  \item \textbf{Model}. A novel statistical model, the multi-layer latent class model (multi-layer LCM), is introduced to capture the latent class structures of multi-layer categorical data with polytomous responses. This model is a multi-layer version of the traditional LCM and it can generate synthetic multi-layer categorical data with known latent classes.
  \item \textbf{Algorithms}. We propose three spectral algorithms for discovering subjects' latent classes and other model parameters within the multi-layer LCM framework. These methods leverage the K-means algorithm on the singular value decomposition or eigen-decomposition of three aggregated matrices: the sum of response matrices, the sum of Gram matrices, and the debiased sum of Gram matrices. We also introduce a method for finding the optimal number of latent classes by integrating the three proposed methods with an averaged modularity metric that assesses the quality of the estimated latent classes in multi-layer categorical data.
  \item \textbf{Theoretical guarantees}. Estimation consistency of our proposed estimation methods by considering data sparsity is built rigorously. The theoretical findings highlight the advantage of multiple layers over a single layer in facilitating the discovery of latent structures and the superiority of the algorithm based on the debiased sum of Gram matrices.
\end{itemize}

\emph{Notation:} The set $\{1,2,\ldots,m\}$ is denoted by $[m]$. The $m\times m$ identity matrix is represented by $I_{m\times m}$. For a matrix $H$, let $H'$ represent its transpose. The notation $H(s,:)$ refers to the submatrix of rows indexed by the set $s$. The Frobenius norm of $H$ is $\|H\|_F$, while the spectral norm is $\|H\|$. Let $\sigma_k(H)$ ($\lambda_k(H)$) stand for the $k$-th largest singular value (eigenvalue in magnitude) of $H$. The rank of $H$ is $\mathrm{rank}(H)$. Expectation and probability are represented by $\mathbb{E}(\cdot)$ and $\mathbb{P}(\cdot)$, respectively.

%The remainder of this paper is structured as follows: Section \ref{sec2} introduces the multi-layer latent class model. Section \ref{sec3} presents three spectral methods. Section \ref{sec4} outlines the estimation consistency for these methods. Section \ref{sec5} offers a method for estimating the number of latent classes. Section \ref{sec6} conducts experimental studies. Section \ref{sec7}  concludes. \ref{SecProofs} presents the proofs.
\section{Model}\label{sec2}
Consider a $L$-layer categorical data with $N$ common subjects and $J$ common items shared across all layers. Let the $N$-by-$J$ matrix $R_{l}$ denote the observed response matrix for the $l$-th layer, where $R_{l}(i,j)\in\{0,1,2,\ldots,M\}$ for $l\in[L], i\in[N], j\in[J]$. Suppose all subjects are partitioned into $K$ disjoint common subgroups $\{\mathcal{C}_{1}, \mathcal{C}_{2}, \ldots, \mathcal{C}_{K}\}$. Define $\mathcal{C}=\bigcup_{k\in[K]}\mathcal{C}_{k}\equiv\{1,2,\ldots,N\}$ as the collection of all subjects. Suppose that the number of subgroups, $K$, is known and significantly smaller than $\mathrm{min}(N, J)$. Let $Z\in\{0,1\}^{N\times K}$ be a matrix where its $(i,k)$-th element is 1 if subject $i$ is in the $k$-th latent class and 0 otherwise, for $i\in[N], k\in[K]$. Since each subject is exclusively assigned to a single subgroup, every row in $Z$ contains precisely one entry that equals 1. In this paper, we refer to $Z$ as the classification matrix. Assuming that there are no empty latent classes, we have $\mathrm{rank}(Z)=K$. Define an $N$-by-$1$ vector $\ell$ such that its $i$-th element is $k$ if $Z(i,k)=1$ for $i\in[N]$. Let $N_{k}$ denote the number of subjects in the $k$-th latent class for $k\in[K]$.

Let the $J$-by-$K$ matrix $\Theta_{l}$ be the item parameter matrix for the $l$-th layer data for $l \in [L]$, where its element are in the interval $[0,M]$, i.e., $\Theta_{l} \in [0,M]^{J \times K}$. Across different layers, the item parameter matrix $\Theta_{l}$ can vary. Our multi-layer latent class model (multi-layer LCM) assumes that the response of subject $i$ to item $j$ in the $l$-th layer of categorical data independently follows a Binomial distribution with probability $\frac{\Theta_{l}(j,\ell(i))}{M}$ and $M$ independent trials. Specifically, for $l \in [L], i \in [N], j \in [J]$, our multi-layer LCM generates $R_{l}(i,j)$ as follows:
\begin{align}\label{LCMBinomial}
\mathbb{P}(R_{l}(i,j)=m)=\frac{M!}{m!(M-m)!}\left(\frac{\Theta_{l}(j,\ell(i))}{M}\right)^{m}\left(1-\frac{\Theta_{l}(j,\ell(i))}{M}\right)^{M-m}, \quad 0\leq m\leq M.
\end{align}
Define $\mathscr{R}_{l}$ as $Z\Theta'_{l}$ and refer to it as the population response matrix for $l \in [L]$. Our multi-layer LCM defined in Equation (\ref{LCMBinomial}) can be simplified as
\begin{align}\label{R0ZTheta}
\mathscr{R}_{l}:=Z\Theta'_{l}, \quad R_{l}(i,j)\sim\mathrm{Binomial}(M,\frac{\mathscr{R}_{l}(i,j)}{M}) \quad \text{for } i \in [N], j \in [J], l \in [L].
\end{align}
From Equation (\ref{R0ZTheta}), we see that our multi-layer LCM is parameterized by $(Z,\{\Theta_{l}\}^{L}_{l=1})$ and it generates the $L$ observed response matrices $\{R_{l}\}^{L}_{l=1}$ through the following steps:

\texttt{Step (a)}: Set $Z \in \{0,1\}^{N \times K}$ and $\Theta_{l} \in [0,M]^{J \times K}$ for $l \in [L]$.

\texttt{Step (b)}: Set $\mathscr{R}_{l} = Z\Theta'_{l}$ for $l \in [L]$.

\texttt{Step (c)}: Let $R_{l}(i,j)$ be a random variable generated from a Binomial distribution, where the probability of success is $\frac{\mathscr{R}_{l}(i,j)}{M}$ and there are $M$ independent trials, for all $i \in [N], j \in [J], l \in [L]$.

Given the $L$ observed response matrices $\{R_{l}\}^{L}_{l=1}$, our objective is to recover the subjects' latent classes and the item parameter matrices $\{\Theta_{l}\}^{L}_{l=1}$. In the subsequent section, we will develop efficient methodologies to accomplish this objective.
\begin{rem}
Our multi-layer LCM is specifically designed to model multi-layer categorical data with binary responses when $M=1$. This implies that for each layer $l$ in $[L]$, the response matrices $R_{l}$ belong to the set $\{0,1\}^{N\times J}$. In the scenario where $L=1$, our multi-layer LCM is simplified to the traditional LCM, which is customized for categorical data with polytomous responses. Additionally, if we set $L=1$ and $M=1$, our model degenerates to LCM designed for categorical data with binary responses.
\end{rem}
\section{Algorithms}\label{sec3}
To design efficient and practical algorithms for estimating the model parameters $(Z,\{\Theta_{l}\}^{L}_{l=1})$ given $\{R_{l}\}^{L}_{l=1}$, suppose that all population response matrices $\{\mathscr{R}_{l}\}^{L}_{l=1}$ are known in advance. Initially, let $\mathscr{R}_{\mathrm{sum}}$ and $\mathcal{S}_{\mathrm{sum}}$ be two aggregation matrices such that $\mathscr{R}_{\mathrm{sum}}=\sum_{l\in[L]}\mathscr{R}_{l}$ and $\mathcal{S}_{\mathrm{sum}}=\sum_{l\in[L]}\mathscr{R}_{l}\mathscr{R}'_{l}$. Note that $\mathscr{R}_{\mathrm{sum}}$ is asymmetric while $\mathcal{S}_{\mathrm{sum}}$ is symmetric. The subsequent lemma lays the groundwork for the proposed estimation procedures.

\begin{lem}\label{SVDEigenDecomposition}
For a multi-layer LCM parameterized by $(Z,\{\Theta_{l}\}^{L}_{l=1})$, for $l\in[L], k\neq \tilde{k}, k\in[K], \tilde{k}\in[K]$, we have the following conclusions:
\begin{enumerate}
  \item $\Theta_{l}=\mathscr{R}'_{l}Z(Z'Z)^{-1}$.
  \item When $\mathrm{rank}(\sum_{l\in[L]}\Theta_{l})=K$, let $\mathscr{R}_{\mathrm{sum}}=U\Lambda B'$ be the compact singular value decomposition (SVD) of $\mathscr{R}_{\mathrm{sum}}$, where $U$ and $B$ satisfy $U'U=I_{K\times K}$ and $B'B=I_{K\times K}$, respectively. Then $U=ZX$ and $X$ satisfies
  \begin{align}\label{XDistance}
  \|X(k,:)-X(\tilde{k},:)\|_{F}=\sqrt{\frac{1}{N_{k}}+\frac{1}{N_{\bar{k}}}},
  \end{align}
  where $X$ is a full rank matrix.
  \item When $\mathrm{rank}(\sum_{l\in[L]}\Theta'_{l}\Theta_{l})=K$, let $\mathcal{S}_{\mathrm{sum}}=V\Sigma V'$ be the compact eigen-decomposition of $\mathcal{S}_{\mathrm{sum}}$, where $V$ satifies $V'V=I_{K\times K}$. Then $V=ZY$ and $Y$ satisfies
  \begin{align}\label{YDistance}
  \|Y(k,:)-Y(\bar{k},:)\|_{F}=\sqrt{\frac{1}{N_{k}}+\frac{1}{N_{\bar{k}}}},
  \end{align}
  where $Y$ is a full rank matrix.
\end{enumerate}
\end{lem}
The 2nd statement of Lemma \ref{SVDEigenDecomposition} implies that running the K-means algorithm to all rows of $U$ with $K$ clusters can recover the classification matrix $Z$ exactly. After recovering $Z$, we can recover $\Theta_{l}$ by setting $\Theta_{l}=\mathscr{R}'_{l}Z(Z'Z)^{-1}$ for $l\in[L]$. Similar arguments hold for $V$.

In practice, the $L$ response matrices $\{R_{l}\}^{L}_{l=1}$ are observed instead of their expectations $\{\mathscr{R}_{l}\}^{L}_{l=1}$. For the real case, we define three aggregation matrices: $R_{\mathrm{sum}}=\sum_{l\in[L]}R_{l}$, $S_{\mathrm{sum}}=\sum_{l\in[L]}R_{l}R'_{l}$, and $\tilde{S}_{\mathrm{sum}}=\sum_{l\in[L]}(R_{l}R'_{l}-D_{l})$, where $R_{\mathrm{sum}}$ is the sum of response matrices, $S_{\mathrm{sum}}$ is the sum of Gram matrices, $\tilde{S}_{\mathrm{sum}}$ is the debiased sum of Gram matrices, and $D_{l}$ is an $N\times N$ diagonal matrix with its $i$-th diagonal entry being $\sum_{j\in[J]}R^{2}_{l}(i,j)$ for $i\in[N], l\in[L]$. It is evident that $\mathbb{E}(R_{\mathrm{sum}})=\mathscr{R}_{\mathrm{sum}}$. By the analysis in \citep{lei2023bias, su2024spectral}, we know that $\tilde{S}_{\mathrm{sum}}$ is a debiased estimator of $\mathcal{S}_{\mathrm{sum}}$, while $S_{\mathrm{sum}}$ is a biased estimator. Let $\hat{U}\hat{\Lambda}\hat{B}'$ be the top $K$ SVD of $R_{\mathrm{sum}}$, where $\hat{U}'\hat{U}=I_{K\times K}$, $\hat{B}'\hat{B}=I_{K\times K}$, and $\hat{\Lambda}$ is a $K\times K$ diagonal matrix with $\hat{\Lambda}(k,k)=\sigma_{k}(R_{\mathrm{sum}})$ for $k\in[K]$. Since the expectation of $R_{\mathrm{sum}}$ is $\mathscr{R}_{\mathrm{sum}}$, it is expected that running K-means to $\hat{U}$'s rows yields a satisfactory estimation of the classification matrix $Z$. Let $\hat{Z}$ be the estimated classification matrix; $\theta_{l}$ can be estimated by setting $R'_{l}\hat{Z}(\hat{Z}'\hat{Z})^{-1}$ for $l\in[L]$. Similarly, let $\hat{V}\hat{\Sigma}\hat{V}'$ be the top $K$ eigen-decomposition of $\tilde{S}_{\mathrm{sum}}$, where $\hat{V}\hat{V}'=I_{K\times K}$ and $\hat{\Sigma}$ is a $K\times K$ diagonal matrix with its $k$-th diagonal entry being  $\lambda_{k}(\tilde{S}_{\mathrm{sum}})$ for $k\in[K]$. Because $\tilde{S}_{\mathrm{sum}}$ is a good estimator of $\mathcal{S}_{\mathrm{sum}}$, applying the K-means algorithm to $\hat{V}$'s rows provides a good estimation of $Z$. Algorithms \ref{alg:Sum} and \ref{alg:DeSC} summarize the above analysis. When we replace $\tilde{S}_{\mathrm{sum}}$ with $S_{\mathrm{sum}}$ in Algorithm \ref{alg:DeSC}, we call the new approach as \textbf{L}atent \textbf{C}lass \textbf{A}nalysis by the \textbf{S}um \textbf{o}f \textbf{G}ram matrices (LCA-SoG).
\begin{algorithm}
\caption{\textbf{L}atent \textbf{C}lass \textbf{A}nalysis by the \textbf{S}um \textbf{o}f \textbf{R}esponse matrices (LCA-SoR)}
\label{alg:Sum}
\begin{algorithmic}[1]
\Require $L$ observed response matrix $\{R_{l}\}^{L}_{l=1}$ and number of latent classes $K$, where $R_{l}\in\{0,1,2,\ldots, M\}^{N\times J}$ for $l\in[L]$.
\Ensure Estimated classification matrix $\hat{Z}$ and $L$ estimated item parameter matrices $\{\hat{\Theta}_{l}\}^{L}_{l=1}$.
\State Set $R_{\mathrm{sum}}=\sum_{l\in[L]}R_{l}$.
\State Compute $\hat{U}\hat{\Lambda}\hat{B}'$, the top $K$ SVD of $R_{\mathrm{sum}}$.
\State Conduct K-means algorithm on $\hat{U}$'s rows with $K$ latent classes to get the estimated classification matrix $\hat{Z}$.
\State Set $\hat{\Theta}_{l}=R'_{l}\hat{Z}(\hat{Z}'\hat{Z})^{-1}$ for $l\in[L]$.
\end{algorithmic}
\end{algorithm}

\begin{algorithm}
\caption{\textbf{L}atent \textbf{C}lass \textbf{A}nalysis by the \textbf{D}ebiased \textbf{S}um \textbf{o}f \textbf{G}ram matrices (LCA-DSoG)}
\label{alg:DeSC}
\begin{algorithmic}[1]
\Require $\{R_{l}\}^{L}_{l=1}$ and $K$.
\Ensure $\hat{Z}$ and $\{\hat{\Theta}_{l}\}^{L}_{l=1}$.
\State Set $\tilde{S}_{\mathrm{sum}}=\sum_{l\in[L]}(R_{l}R'_{l}-D_{l})$, where the $i$-th entry of the diagonal matrix $D_{l}$ is $D_{l}(i,i)=\sum_{j\in[J]}R^{2}_{l}(i,j)$ for $i\in[N], l\in[L]$.
\State Obtain $\hat{V}\hat{\Sigma}\hat{V}'$, the top $K$ eigen-decomposition of $\tilde{S}_{\mathrm{sum}}$.
\State Conduct K-means algorithm on $\hat{V}$'s rows with $K$ latent classes to get $\hat{Z}$.
\State Set $\hat{\Theta}_{l}=R'_{l}\hat{Z}(\hat{Z}'\hat{Z})^{-1}$ for $l\in[L]$.
\end{algorithmic}
\end{algorithm}
\begin{rem}
It is worth mentioning that our debiased sum of Garm matrices  $\tilde{S}_{\mathrm{sum}}$ takes the form $\sum_{l\in[L]}(R_{l}R'_{l}-D_{l})$. Here, $D_{l}(i, i)$ is computed as $\sum_{j\in[J]}R^{2}_{l}(i,j)$, rather than $\sum_{j\in[J]}R_{l}(i,j)$ as stated in \citep{su2024spectral} for multi-layer directed unweighted networks. This adjustment is made to accommodate the polytomous responses. The rationale behind using $\sum_{j\in[J]}R^{2}_{l}(i,j)$ for calculating $D_{l}(i,i)$ can be further elaborated in the proof of bounding $\|\tilde{S}_{\mathrm{sum}}-\mathcal{S}_{\mathrm{sum}}\|$ presented in Lemma \ref{boundSumDeSumMLLCM}.
\end{rem}
\section{Main results}\label{sec4}
This section presents theoretical guarantees for LCA-SoR, LCA-DSoG, and LCA-SoG under the multi-layer LCM parameterized by $(Z,\{\Theta_{l}\}^{L}_{l=1})$. Recall that for $l\in[L]$, $\Theta_{l}\in[0,M]^{J\times K}$ does not imply that the maximum element of $\Theta_{l}$ is $M$. Instead, if we define $\rho=\mathrm{max}_{j\in[J], k\in[K],l\in[L]}\Theta_{l}(j,k)$, then $\rho\in(0,M]$. Equation (\ref{LCMBinomial}) indicates that $\mathbb{P}(R_{l}(i,j)=0)=(1-\frac{\Theta_{l}(j,\ell(i))}{M})^{M}$, suggesting that decreasing $\rho$ increases the probability of generating a no-response (0). Therefore, $\rho$ governs the general sparsity of multi-layer categorical data., and we term it the sparsity parameter. Generally, recovering latent classes and item parameter matrices becomes challenging when multi-layer categorical data contains numerous no-responses. Hence, it is crucial to thoroughly examine the influence of the sparsity parameter $\rho$ on the effectiveness of the proposed methodologies. For convenience, set $B_{l}=\frac{\Theta_{l}}{\rho}$ for $l\in[L]$ and $\mathcal{B}=\{B_{l}\}^{L}_{l=1}$. We see that $B_{l}\in[0,1]^{J\times K}$ for $l\in[L]$. Assumptions \ref{Assum1} and \ref{Assum2} establish the minimum prerequisites for the sparsity parameter $\rho$ in building the theoretical guarantees of LCA-SoR and LCA-DSoG (LCA-SoG), respectively. These assumptions suggest that as the number of layers $L$ increases, the minimum sparsity requirement for the data decreases accordingly.
\begin{assum}\label{Assum1}
(LCA-SoR's requirement on sparsity) $\rho L\mathrm{max}(N,J)\gg M^{2}\mathrm{log}(N+J+L)$.
\end{assum}
\begin{assum}\label{Assum2}
(LCA-DSoG's and LCA-SoG's requirements on sparsity) $\rho^{2}NJL\gg M^{4}\mathrm{log}(N+J+L)$.
\end{assum}
Assumptions \ref{Assum11} and \ref{Assum22} outline the necessary conditions for $\mathcal{B}$ in the theoretical analysis of LCA-SoR and LCA-DSoG (LCA-SoG), respectively. Both assumptions are mild due to the fact that $\sum_{l\in[L]}B_{l}$ and $\sum_{l\in[L]}B'_{l}B_{l}$ represent the summation of $L$ matrices, where $B_{l}\in[0,1]^{J\times K}$ for $l\in[L]$, and $K$ is significantly smaller than $J$.
\begin{assum}\label{Assum11}
(LCA-SoR's requirement on $\mathcal{B}$) $\sigma_{K}(\sum_{l\in[L]}B_{l})\geq a_{1}\sqrt{J}L$ for some positive constant $a_{1}>0$.
\end{assum}
\begin{assum}\label{Assum22}
(LCA-DSoG's and LCA-SoG's requirements on $\mathcal{B}$) $\sigma_{K}(\sum_{l\in[L]}B'_{l}B_{l})\geq a_{2}JL$ for some positive constant $a_{2}>0$.
\end{assum}
Lemma \ref{boundSumDeSumMLLCM} plays a crucial role in our main results, as it characterizes the spectral norm of $R_{\mathrm{sum}}-\mathscr{R}_{\mathrm{sum}}$, $\tilde{S}_{\mathrm{sum}}-\mathcal{S}_{\mathrm{sum}}$, and $S_{\mathrm{sum}}-\mathcal{S}_{\mathrm{sum}}$.
\begin{lem}\label{boundSumDeSumMLLCM}
For a multi-layer LCM parameterized by $(Z,\{\Theta_{l}\}^{L}_{l=1})$, with probability at least $1-o((J+N+L)^{-3})$,
\begin{itemize}
  \item suppose that Assumption \ref{Assum1} is satisfied, we have
\begin{align*}
      \|R_{\mathrm{sum}}-\mathscr{R}_{\mathrm{sum}}\|=O(\sqrt{\rho L\mathrm{max}(J,N)\mathrm{log}(J+N+L)}).
\end{align*}
  \item suppose that Assumption \ref{Assum2} is satisfied, we have
\begin{align*}
\|\tilde{S}_{\mathrm{sum}}-\mathcal{S}_{\mathrm{sum}}\|=O(\sqrt{\rho^{2}JNL\mathrm{log}(N+J+L)})+\rho^{2}JL\mathrm{~and~}\|S_{\mathrm{sum}}-\mathcal{S}_{\mathrm{sum}}\|=O(\sqrt{\rho^{2}JNL\mathrm{log}(J+N+L)})+(\rho^{2}+M^{2})JL.
\end{align*}
\end{itemize}
\end{lem}
For $k\in[K]$, let $\hat{\mathcal{C}}_{k}$ denote the collection of subjects assigned to the $k$-th estimated subgroup. Specifically, for $i\in[N]$, subject $i$ is a member of $\hat{\mathcal{C}}_{k}$ if $\hat{Z}(i,k)=1$. Define $\hat{\mathcal{C}} = \cup_{k=1}^{K} \hat{\mathcal{C}}_{k}$ as the aggregate of all estimated latent classes. To quantify the discrepancy between the true partition $\mathcal{C}$ and its estimated version $\hat{\mathcal{C}}$, this paper employs the \emph{Clustering error} metric introduced in Equation (6) of \cite{joseph2016impact}. The formal definition of this metric is provided below:
\begin{align}\label{clusteringerror}
\hat{f} = \min_{p \in \mathcal{P}_{K}} \max_{k \in \{1, \ldots, K\}} \frac{|\mathcal{C}_{k} \cap \hat{\mathcal{C}}_{p(k)}^{c}| + |\mathcal{C}_{k}^{c} \cap \hat{\mathcal{C}}_{p(k)}|}{N_{K}},
\end{align}
where $c$ represents the complementary set and $\mathcal{P}_{K}$ represents the set of all permutations of $\{1, 2, \ldots, K\}$. Furthermore, let $\hat{f}_{LCA-SoR}$, $\hat{f}_{LCA-DSoG}$, and $\hat{f}_{LCA-SoG}$ represent the Clustering errors computed using Equation (\ref{clusteringerror}) for LCA-SoR, LCA-DSoG, and LCA-SoG, respectively, based on the estimated classification matrix $\hat{Z}$ returned by each method. The following Theorem \ref{mainMLLCM} represents the main theoretical result of this paper, offering rigorous upper bounds of error rates of the three proposed methods.
\begin{thm}\label{mainMLLCM}
For a multi-layer LCM parameterized by $(Z,\{\Theta_{l}\}^{L}_{l=1})$, with probability at least $1-o((J+N+L)^{-3})$,
\begin{itemize}
\item suppose that Assumptions \ref{Assum1} and \ref{Assum11} are satisfied, we have
\begin{align*}
\hat{f}_{LCA-SoR}=O(\frac{K^{2}N_{\mathrm{max}}\mathrm{max}(J,N)\mathrm{log}(J+N+L)}{\rho N^{2}_{\mathrm{min}}JL}),
\end{align*}
where $N_{\mathrm{max}}=\mathrm{max}_{k\in[K]}N_{k}$ and $N_{\mathrm{min}}=\mathrm{min}_{k\in[K]}N_{k}$.
\item suppose that Assumptions \ref{Assum2} and \ref{Assum22} are satisfied, we have
\begin{align*}
\hat{f}_{LCA-DSoG}=O(\frac{K^{2}N_{\mathrm{max}}N\mathrm{log}(J+N+L)}{\rho^{2}N^{3}_{\mathrm{min}}JL})+O(\frac{K^{2}N_{\mathrm{max}}}{N^{3}_{\mathrm{min}}}) \mathrm{~and~}\hat{f}_{LCA-SoG}=O(\frac{K^{2}N_{\mathrm{max}}N\mathrm{log}(J+N+L)}{\rho^{2}N^{3}_{\mathrm{min}}JL})+O(\frac{K^{2}N_{\mathrm{max}}M^{4}}{N^{3}_{\mathrm{min}}\rho^{4}}).
\end{align*}
\item for all methods, suppose that Assumptions \ref{Assum1} and \ref{Assum11} are satisfied, we have
\begin{align*}
\frac{\|\sum_{l\in[L]}(\hat{\Theta}_{l}-\Theta_{l})\|_{F}}{\|\sum_{l\in[L]}\Theta_{l}\|_{F}}=O(\sqrt{\frac{K\mathrm{max}(J,N)\mathrm{log}(J+N+L)}{\rho N_{\mathrm{min}}JL}}).
\end{align*}
\end{itemize}
\end{thm}
According to Theorem \ref{mainMLLCM}, increasing the sparsity parameter $\rho$ enhances the estimation accuracies of both latent classes and item parameter matrices for all methods. Furthermore, Theorem \ref{mainMLLCM} indicates that increasing $L$ also improves estimation accuracies for all algorithms,  highlighting the significance of considering multiple layers in latent class analysis. Additionally, Theorem \ref{mainMLLCM} suggests that the three proposed methods exhibit no significant difference in estimating the item parameter matrices.

Through the proof of the error bounds for LCA-DSoG and LCA-SoG, we know that $\hat{f}_{LCA-DSoG}$ is consistently smaller than $\hat{f}_{LCA-SoG}$, highlighting the advantages of eliminating $\sum_{l\in[L]}D_{l}$ from $S_{\mathrm{sum}}$. The subsequent lemma compares the error bounds of LCA-DSoG and LCA-SoR, revealing that LCA-DSoG significantly outperforms LCA-SoR when the number of layers $L$ is not excessively large. Additionally,  the error rates are small when $L$ is very large for all methods based on Theorem \ref{mainMLLCM}, indicating no significant difference in performances between these methods for a large $L$. Based on the findings presented in Lemma \ref{CompareSoRDSoG}, and considering that the number of layers $L$ is generally not excessively large in practical applications, we can confidently conclude that LCA-DSoG consistently exhibits superior performance compared to the other two methods.
\begin{lem}\label{CompareSoRDSoG}
Under the same conditions of Theorem \ref{mainMLLCM}, LCA-DSoG significantly outperforms LCA-SoR (i.e., $\hat{f}_{LCA-DSoG}\ll\hat{f}_{LCA-SoR}$) when $L\ll\frac{N^{2}_{\mathrm{min}}\mathrm{max}(J^{2},N^{2})\mathrm{log}(J+N+L)}{JN}$.
\end{lem}
The following corollary simplifies the results in Theorem \ref{mainMLLCM}.
\begin{cor}\label{Corollary}
Under the same conditions of Theorem \ref{mainMLLCM}, suppose that $K=O(1), N_{\mathrm{min}}=O(\frac{N}{K}), N_{\mathrm{max}}=O(\frac{N}{K})$, and $J=O(N)$, we have
\begin{align*}
&\hat{f}_{LCA-SoR}=O(\frac{\mathrm{log}(J+N+L)}{\rho NL}), \hat{f}_{LCA-DSoG}=O(\frac{\mathrm{log}(J+N+L)}{\rho^{2}N^{2}L})+O(\frac{1}{N^{2}}),\\ &\hat{f}_{LCA-SoG}=O(\frac{\mathrm{log}(J+N+L)}{\rho^{2}N^{2}L})+O(\frac{M^{4}}{N^{2}\rho^{4}}), \frac{\|\sum_{l\in[L]}(\hat{\Theta}_{l}-\Theta_{l})\|_{F}}{\|\sum_{l\in[L]}\Theta_{l}\|_{F}}=O(\sqrt{\frac{\mathrm{log}(J+N+L)}{\rho NL}}).
\end{align*}
\end{cor}
According to Corollary \ref{Corollary}, increasing $N$ leads to a decrease in error rates across all methods, thereby emphasizing the significance of incorporating a larger sample size in latent class analysis. According to Corollary \ref{Corollary}, we notice that $\rho$ must decrease at a rate of at least $\frac{\mathrm{log}(N+J+L)}{NL}$ to guarantee adequately low error rates for LCA-SoR. This insight aligns with Assumption \ref{Assum1}. Similar arguments hold for LCA-DSoG and LCA-SoG as well.
\section{Estimating the number of latent classes}\label{sec5}
The true latent classes are typically unavailable for real data, while estimated latent subgroups can always be obtained by applying our methods to the data. Therefore, it is crucial to develop a metric for measuring the quality of estimated latent classes in multi-layer categorical data with polytomous responses. Additionally, for real data, we usually do not know the number of subgroups $K$, making its determination important. For $i\in[N], l\in[L]$, let $A_{l}=R_{l}R'_{l}$ and $d_{l}$ be a diagonal matrix with its $i$-th diagonal entry being $d_{l}(i, i)=\sum_{\bar{i}\in[N]}A_{l}(i,\bar{i})$, where $A_{l}$ forms an assortative weighted network in the literature of network science \citep{newman2003mixing}. Further insights into $A_{l}$ can be found in \citep{qing2023latent}. To quantify the quality of class partitions in multi-layer categorical data, we use the averaged modularity metric proposed in \cite{paul2021null}, which is computed using $\{A_{l}\}^{L}_{l=1}$. Let $\omega_{l}=\frac{1}{2}\sum_{i=1}^{N}d_{l}(i,i)$ for $l\in[L]$. Assuming there are $k$ latent classes and $\hat{Z}\in\{0,1\}^{N\times k}$ is an estimated classification matrix returned by running a certain algorithm $\mathcal{M}$ on multi-layer categorical data described by $\{R_{l}\}^{L}_{l=1}$, the averaged modularity can be calculated as follows:
\begin{align}\label{Modularity}
Q_{\mathcal{M}}(k)=\frac{1}{L}\sum_{l\in[L]}\sum_{i\in[N]}\sum_{\bar{i}\in[N]}\frac{1}{2\omega_{l}}(A_{l}(i,\bar{i})-\frac{d_{l}(i,i)d_{l}(\bar{i},\bar{i})}{2\omega_{l}})\hat{Z}(i,:)\hat{Z}'(\bar{i},:).
\end{align}

When $L=1$, the averaged modularity simplifies to the well-known Newman-Girvan modularity \citep{newman2004finding, newman2006modularity}. A higher value of the averaged modularity signifies superior class partition quality \citep{newman2006modularity, paul2021null}. Hence, for any given method $\mathcal{M}$, it is ideal to have an optimal number of latent classes that maximize this metric. To find the optimal number of subgroups $K$, we choose the $k$ that maximizes $Q_{\mathcal{M}}(k)$ for method $\mathcal{M}$, as suggested in \cite{nepusz2008fuzzy}. When we use method $\mathcal{M}$ to estimate $K$ by maximizing $Q_{\mathcal{M}}(k)$, we conveniently call it K$\mathcal{M}$.
\section{Experimental studies}\label{sec6}
We provide experimental results to illustrate the efficacy of the proposed methodologies in this section. We compare our methods with the following three baseline methods:
\begin{itemize}
  \item \texttt{LCA-SoRK}: This method estimates $Z$ by directly running K-means to $R_{\mathrm{sum}}$ with $K$ subgroups, and estimates $\Theta_{l}$ using the 4th step of Algorithm \ref{alg:Sum} for each $l\in[L]$.
  \item \texttt{LCA-SoGK}: This method estimates $Z$ by directly applying K-means to $S_{\mathrm{sum}}$ and estimates the item parameter matrices using the 4th step of Algorithm \ref{alg:Sum}.
  \item \texttt{LCA-DSoGK}: This method estimates $Z$ by directly applying K-means to $\tilde{S}_{\mathrm{sum}}$ and estimates $\{\Theta_{l}\}^{L}_{l=1}$ in a manner similar to other methods.
\end{itemize}

To quantify the difference between $Z$ and its estimated version $\hat{Z}$, we use not only the \texttt{Clustering error} defined in Equation (\ref{clusteringerror}), but also several other metrics: \texttt{Hamming error} \citep{SCORE}, \texttt{NMI} \citep{strehl2002cluster, danon2005comparing, bagrow2008evaluating, luo2017community}, and \texttt{ARI} \citep{luo2017community, hubert1985comparing, vinh2009information}. See \citep{qing2023community} for detailed information on these metrics. It is worth noting that lower Clustering and Hamming errors indicate better performance, whereas higher NMI and ARI values signify superior performance.

To assess the difference between the true item parameter matrices $\{\Theta_{l}\}^{L}_{l=1}$ and their estimates $\{\hat{\Theta}_{l}\}^{L}_{l=1}$, we employ the \texttt{Relative $l_{2}$ error}, which is defined as $\frac{\|\sum_{l\in[L]}(\hat{\Theta_{l}}-\Theta_{l})\|_{F}}{\|\sum_{l\in[L]}\Theta_{l}\|_{F}}$. This metric is also inversely related to performance, with lower values indicating better performance.

To assess the ability of our methods in estimating $K$, we utilize the \texttt{Accuracy rate} defined as the proportion of instances where a method accurately determines the value of $K$. This metric provides a direct measure of the performance of all $K\mathcal{M}$ methods in the task of estimating $K$.
\begin{figure}
\centering
\resizebox{\columnwidth}{!}{
\subfigure[]{\includegraphics[width=0.2\textwidth]{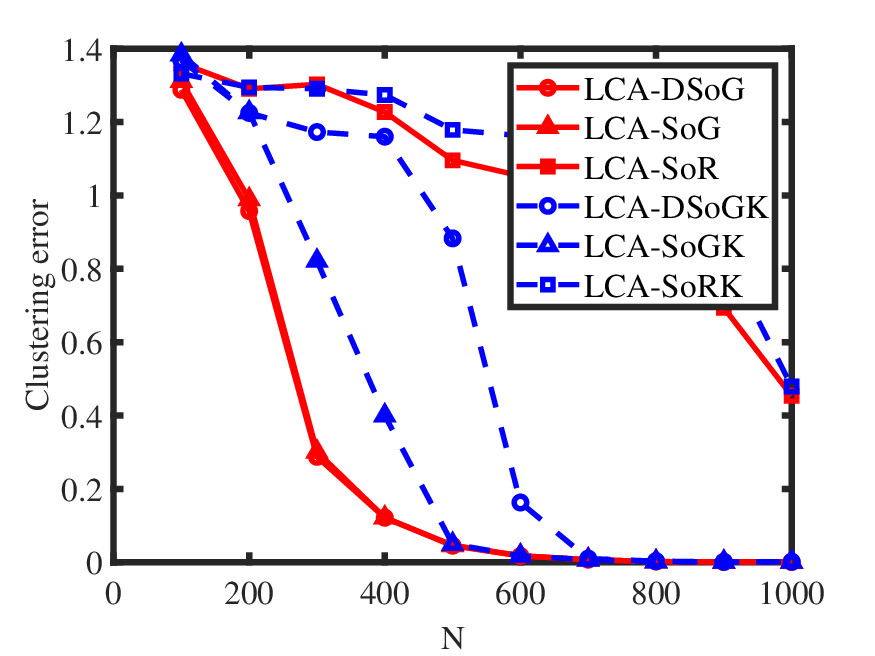}}
\subfigure[]{\includegraphics[width=0.2\textwidth]{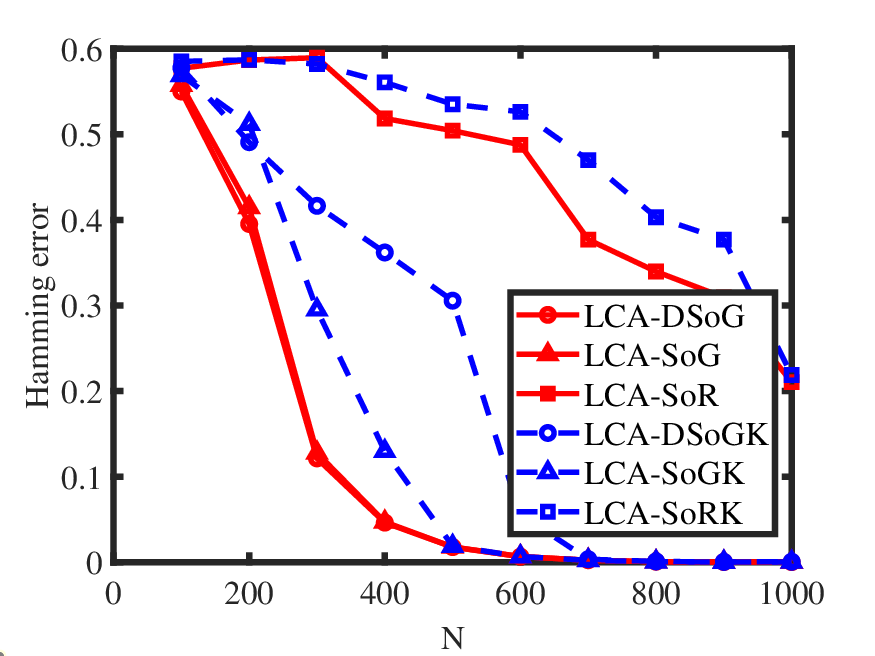}}
}
\resizebox{\columnwidth}{!}{
\subfigure[]{\includegraphics[width=0.2\textwidth]{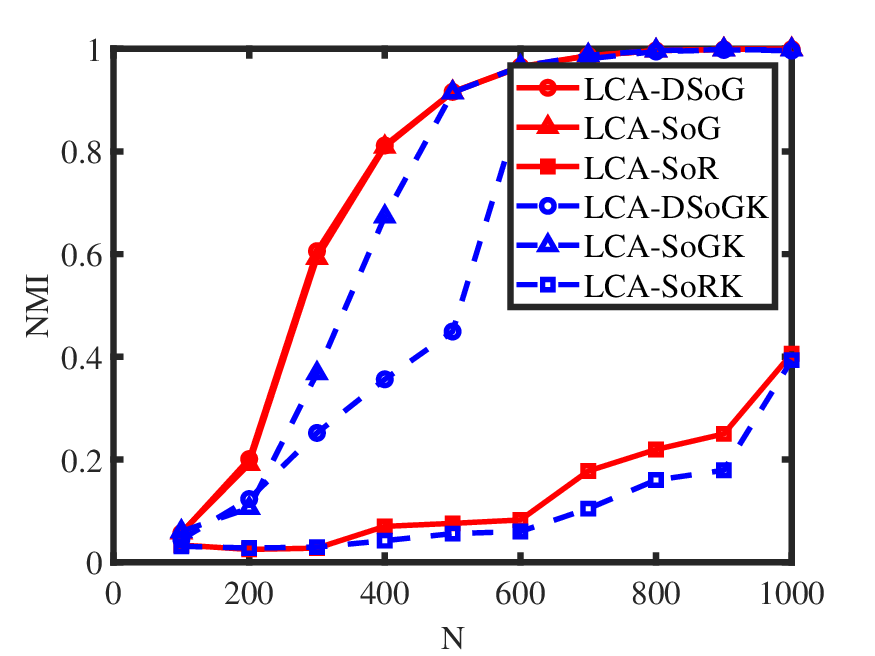}}
\subfigure[]{\includegraphics[width=0.2\textwidth]{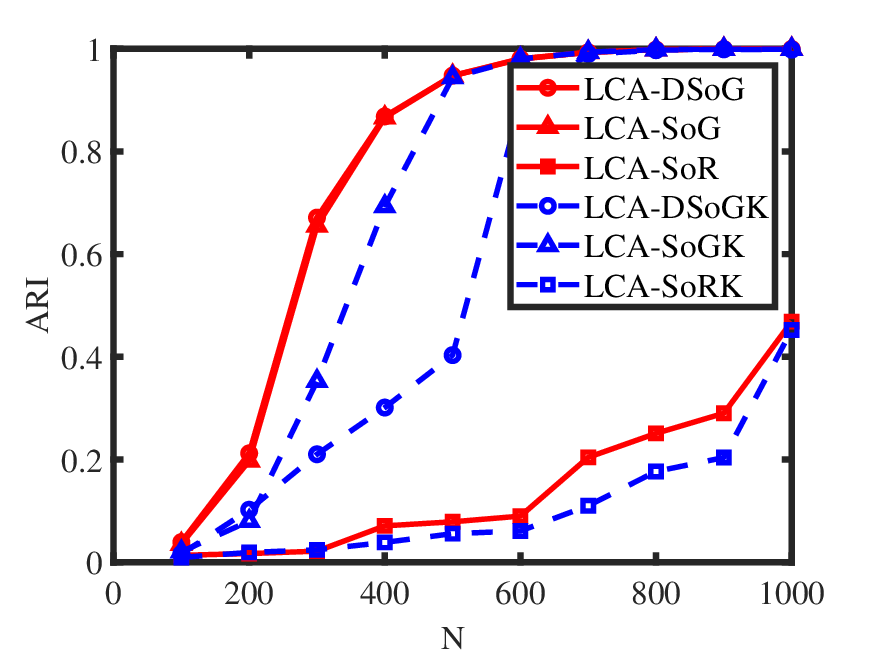}}
}
\resizebox{\columnwidth}{!}{
\subfigure[]{\includegraphics[width=0.2\textwidth]{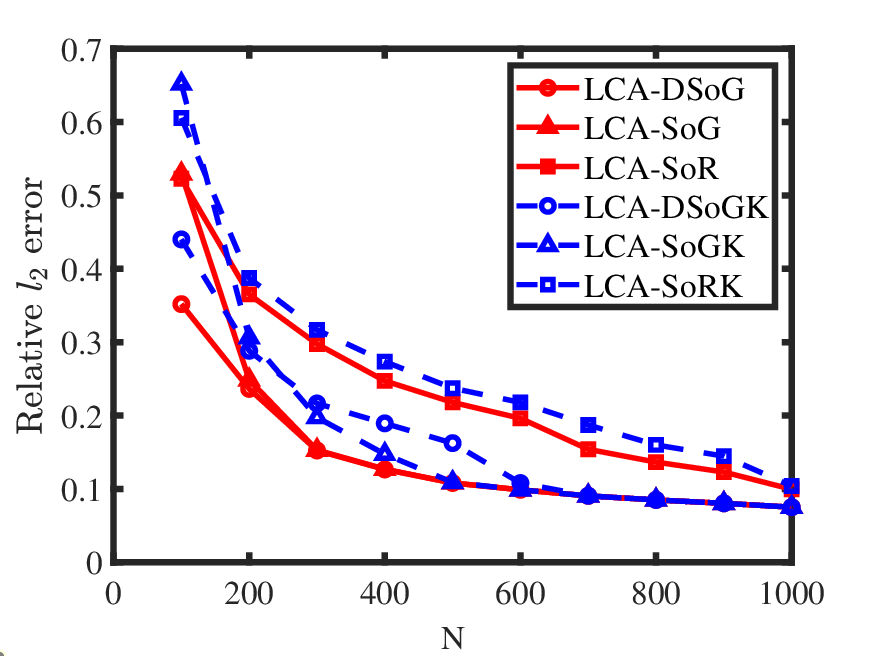}}
\subfigure[]{\includegraphics[width=0.2\textwidth]{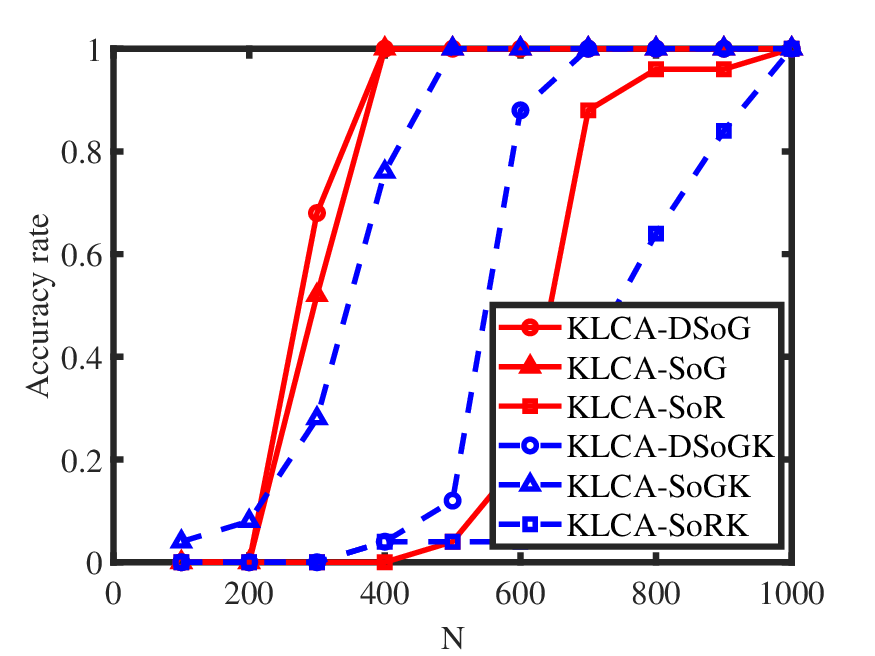}}
}
\caption{Numerical results of Experiment 1.}
\label{Ex1} %% label for entire figure
\end{figure}
\begin{figure}
\centering
\resizebox{\columnwidth}{!}{
\subfigure[]{\includegraphics[width=0.2\textwidth]{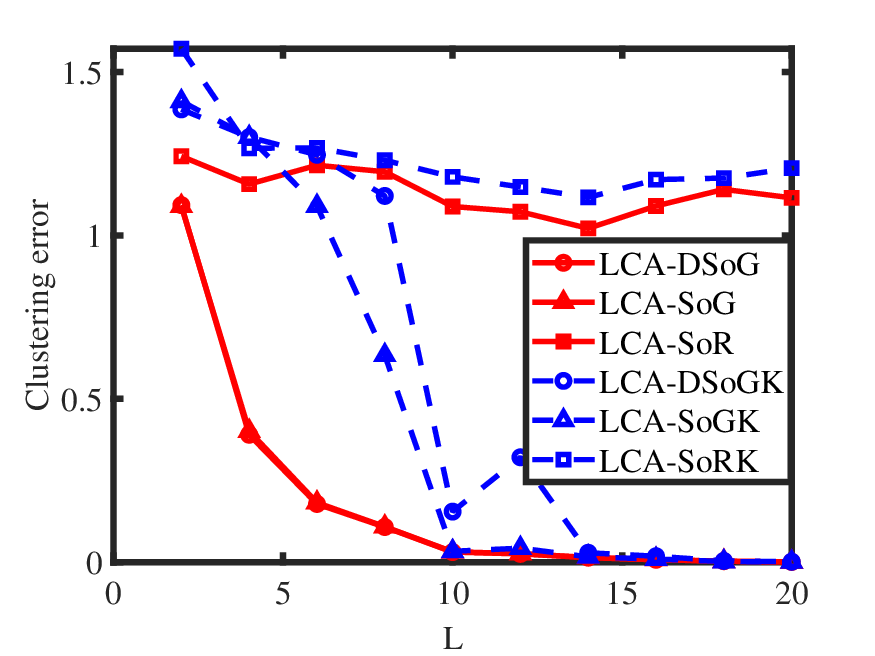}}
\subfigure[]{\includegraphics[width=0.2\textwidth]{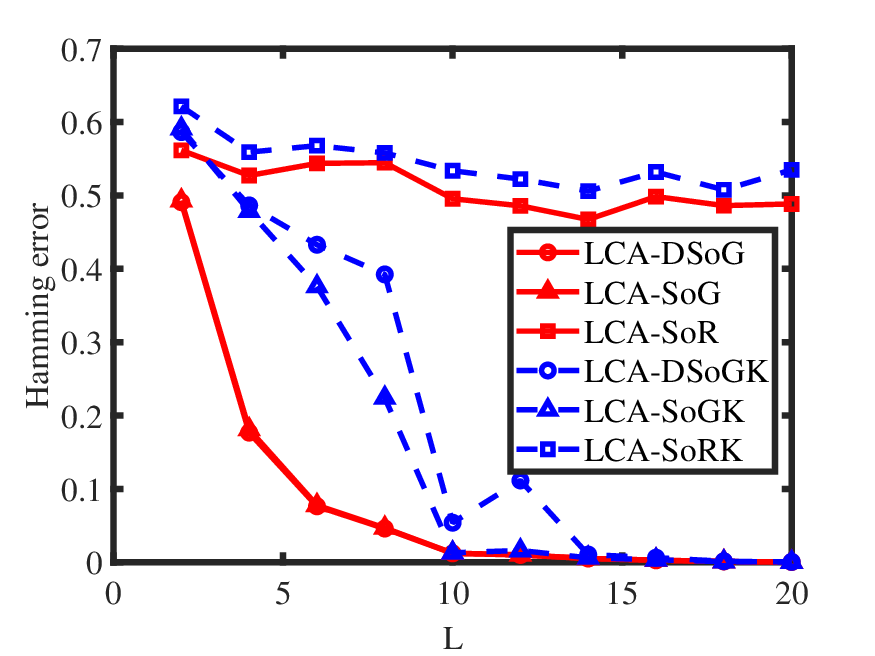}}
}
\resizebox{\columnwidth}{!}{
\subfigure[]{\includegraphics[width=0.2\textwidth]{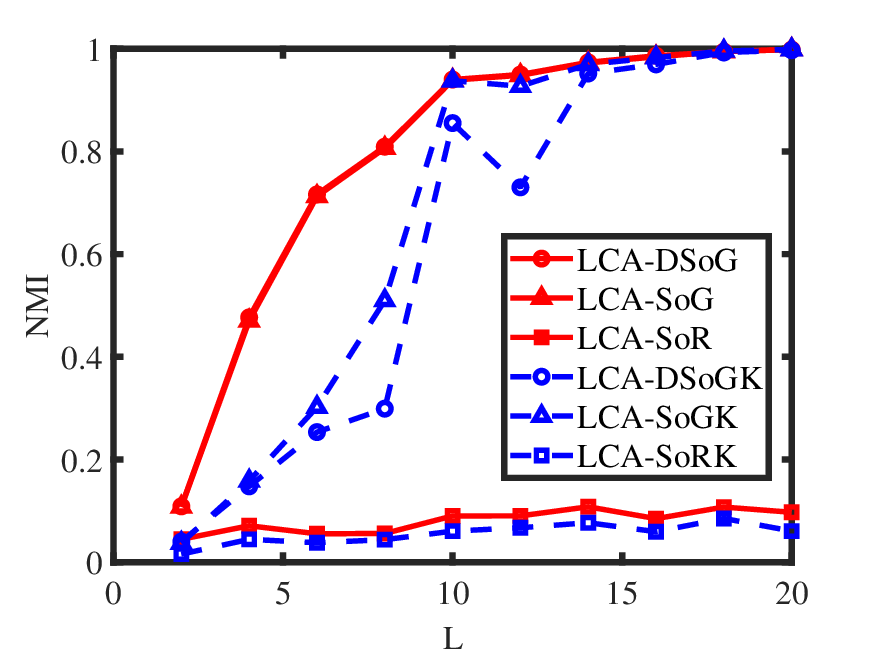}}
\subfigure[]{\includegraphics[width=0.2\textwidth]{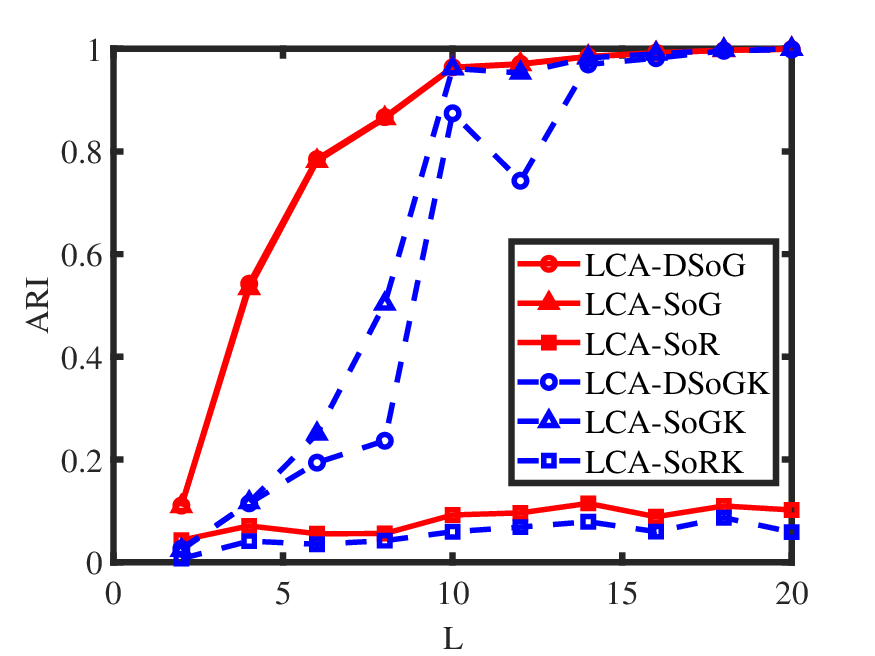}}
}
\resizebox{\columnwidth}{!}{
\subfigure[]{\includegraphics[width=0.2\textwidth]{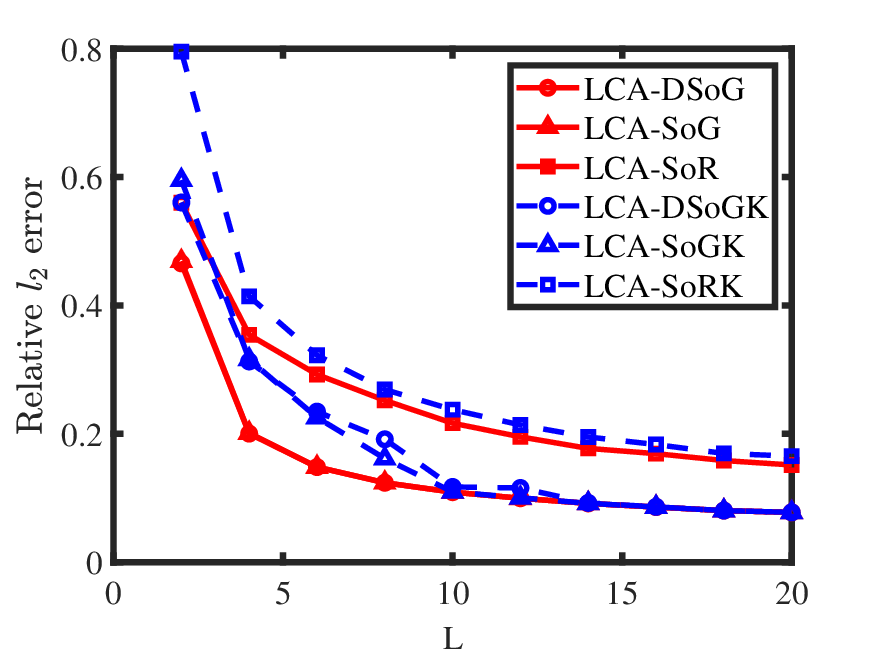}}
\subfigure[]{\includegraphics[width=0.2\textwidth]{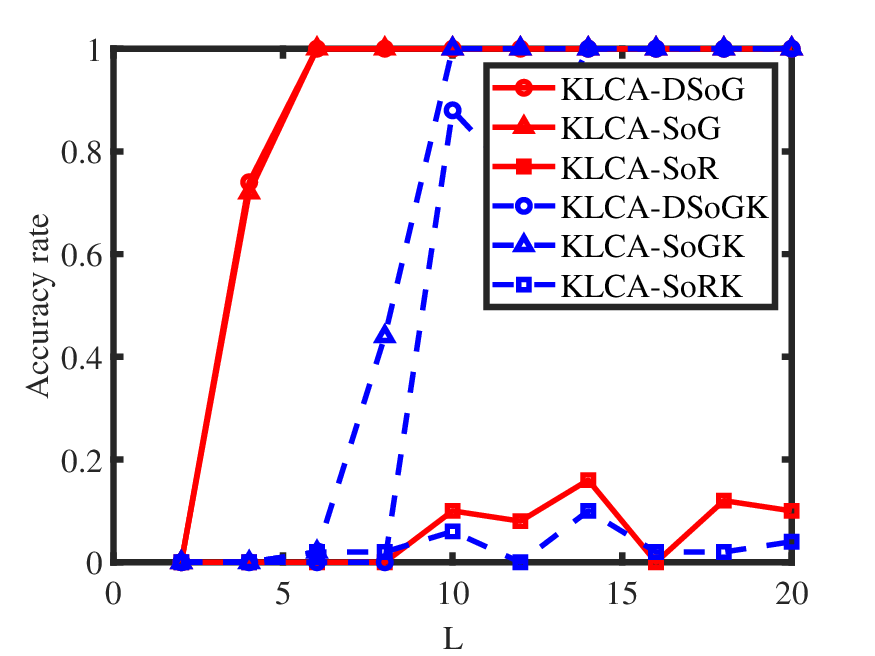}}
}
\caption{Numerical results of Experiment 2.}
\label{Ex2} %% label for entire figure
\end{figure}
\begin{figure}
\centering
\resizebox{\columnwidth}{!}{
\subfigure[]{\includegraphics[width=0.2\textwidth]{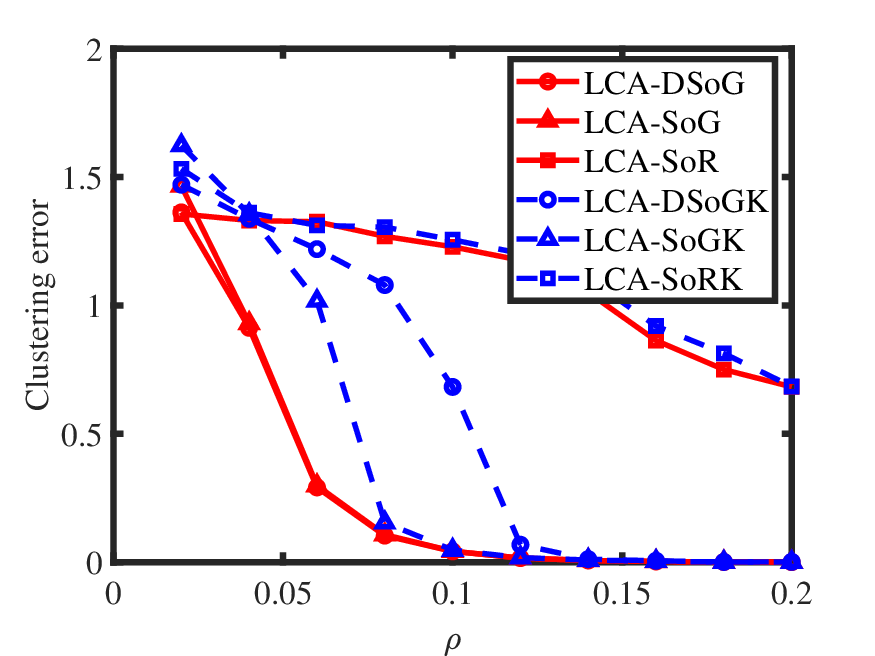}}
\subfigure[]{\includegraphics[width=0.2\textwidth]{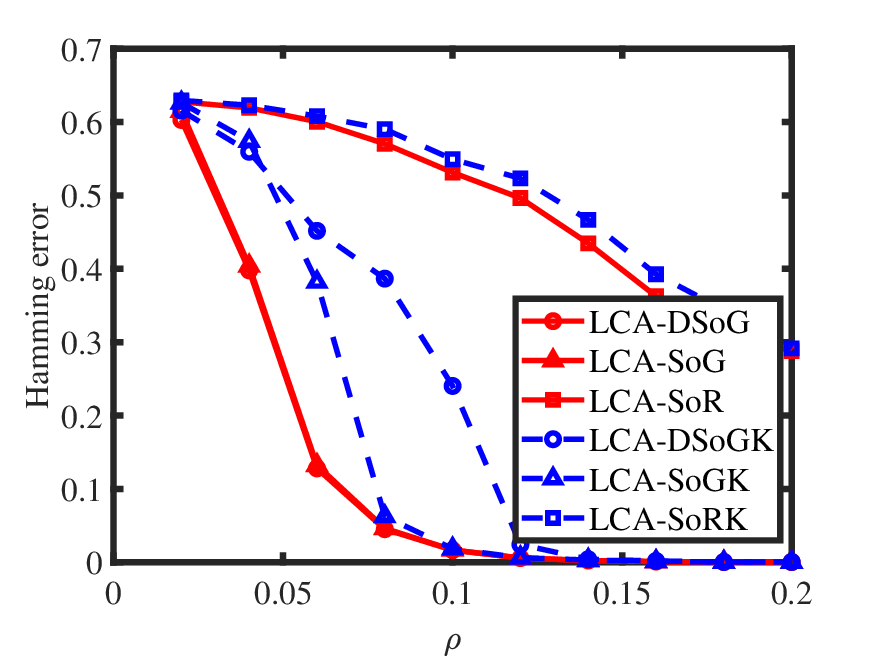}}
}
\resizebox{\columnwidth}{!}{
\subfigure[]{\includegraphics[width=0.2\textwidth]{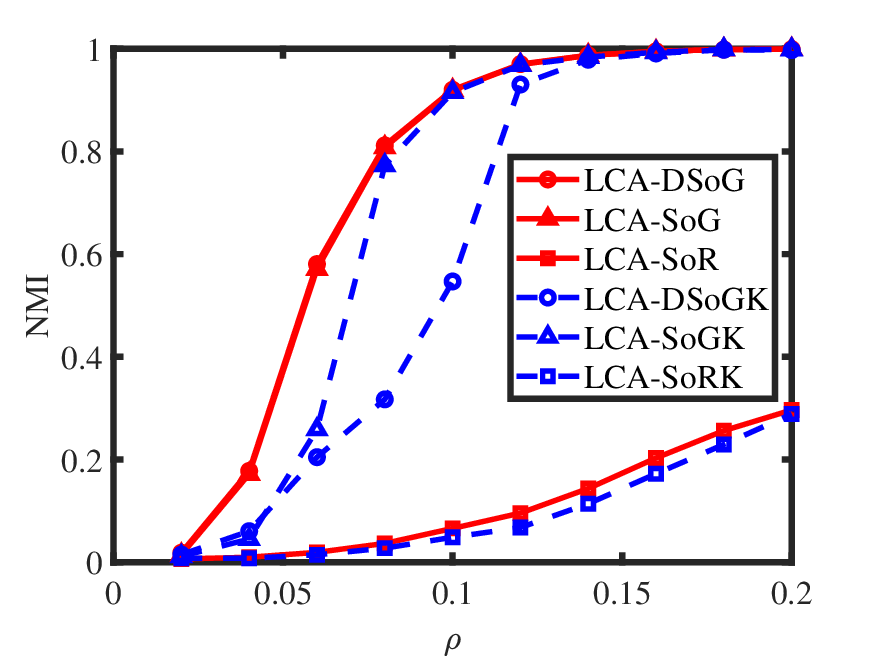}}
\subfigure[]{\includegraphics[width=0.2\textwidth]{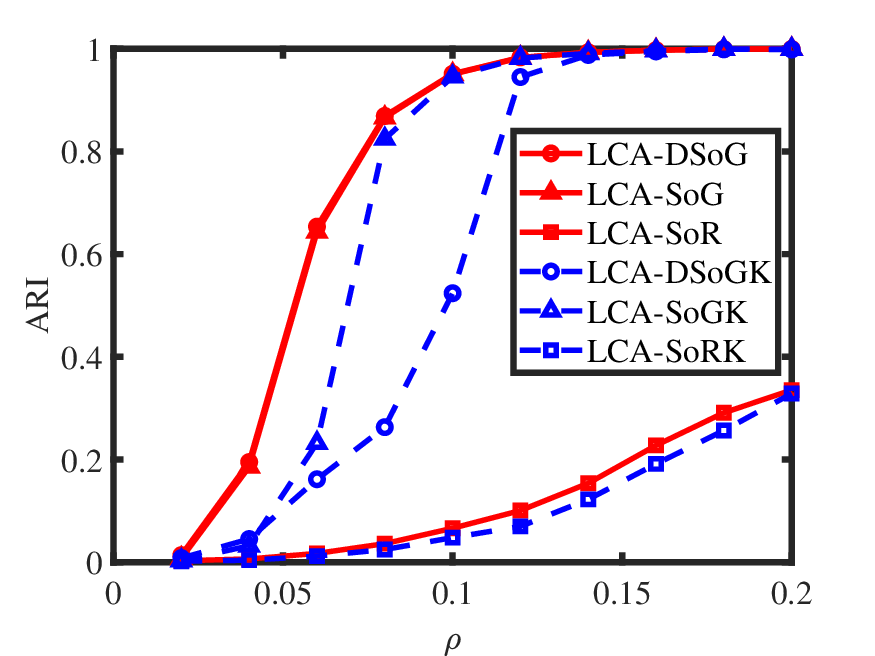}}
}
\resizebox{\columnwidth}{!}{
\subfigure[]{\includegraphics[width=0.2\textwidth]{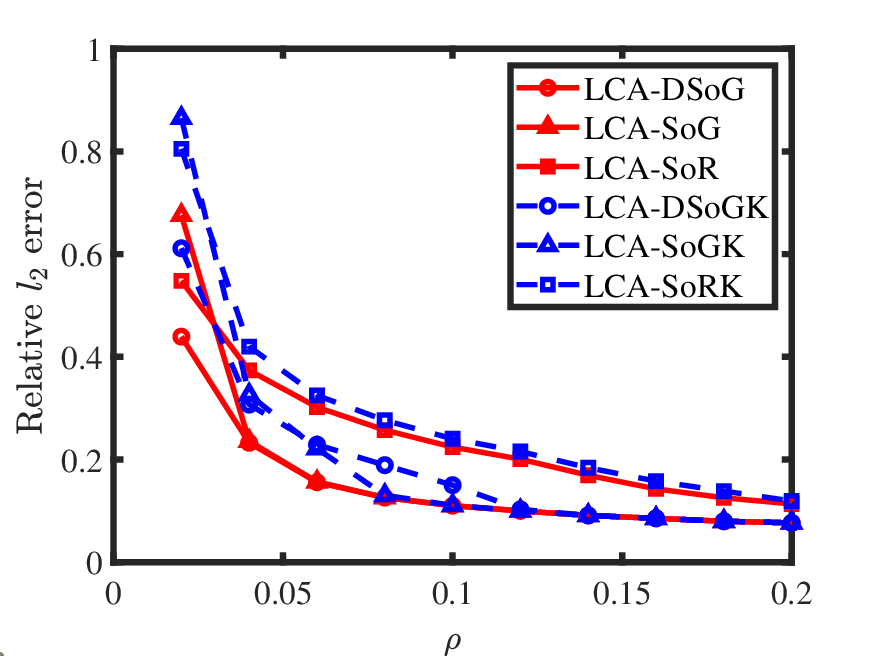}}
\subfigure[]{\includegraphics[width=0.2\textwidth]{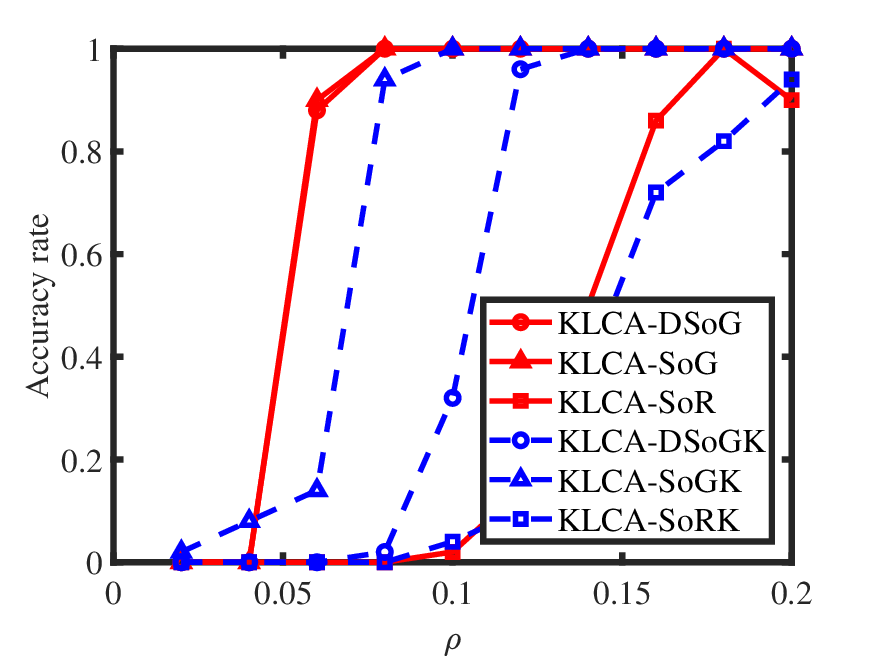}}
}
\caption{Numerical results of Experiment 3.}
\label{Ex3} %% label for entire figure
\end{figure}

In our experiments, we fix $J=\frac{N}{5}, K=3$, and $M=5$. We generate the classification matrix $Z$ by assigning each subject to each latent class with equal probability. Each element of $B_{l}$ is a random value from $(0,1)$ for $l$ in $[L]$. For each experiment, the sparsity parameter $\rho$, the number of layers $L$, and the number of subjects $N$ are independently determined. To ensure statistical significance, we conduct 50 repetitions for every set of parameters and present the average results for each metric.

\emph{Experiment 1: Varying $N$.} Set $L=10, \rho=0.1$, and let $N$ vary within the set $\{100, 200, \ldots, 1000\}$. The results are displayed in Figure \ref{Ex1}. It is easy to observe that (a) All approaches behave better as the number of subjects $N$ increases, which matches our theoretical findings. (b) Our LCA-DSoG slightly outperforms LCA-SoG, and both methods enjoy better performances than LCA-SoR, LCA-DSoGK, LCA-SoGK, and LAC-SoRK. (c) Our KLCA-DSoG and KLCA-SoG perform better than the other four methods in estimating $K$, and they have high accuracy rates when $N$ is large.

\emph{Experiment 2: Varying $L$.} Set $N=500, \rho=0.1$, and let $L$ vary within the set $\{2, 4, \ldots, 20\}$. The results presented in Figure \ref{Ex2} indicate several key insights: (a) Increasing the number of layers $L$ consistently enhances the performance of all algorithms, thereby supporting our theoretical claims and emphasizing the merits of multi-layer categorical data over single-layer categorical data. (b) Our  LCA-DSoG and LCA-SoG algorithms exhibit comparable performances, significantly surpassing their competitors. (c) Our KLCA-DSoG and KLCA-SoG approaches exhibit satisfactory performance in estimating $K$, particularly for larger values of $L$.

\emph{Experiment 3: Varying $\rho$.} Set $N=500, L=10$, and let $\rho$ vary within the set $\{0.02, 0.04, \ldots, 0.2\}$. As displayed in Figure \ref{Ex3}, the following insights emerge: (a) A notable trend is evident in which the performance of all approaches is enhanced with an increase in the sparsity parameter $\rho$, corroborating the theoretical predictions. (b) Our proposed methods, LCA-DSoG and LCA-SoG, exhibit superior performance in estimating latent classes and item parameter matrices, outperforming the competitors for these tasks. (c) For larger values of $\rho$, our KLCA-DSoG and KLCA-SoG methods demonstrate remarkable accuracy in estimating the number of latent classes.
\section{Conclusion and future work}\label{sec7}
We have introduced a novel statistical model, the multi-layer latent class model, specifically designed for multi-layer categorical data with polytomous responses. We also developed three spectral methods—LCA-SoR, LCA-DSoG, and LCA-SoG to fit the model and present practitioners with robust, powerful, and efficient tools for inferring latent classes and item parameter matrices for multi-layer categorical data. Our theoretical results underscore the superiority of leveraging multi-layer categorical data over single-layer categorical data, as well as the superiority of LCA-DSoG over LCA-SoR and LCA-SoG, emphasizing the benefits of incorporating debiased spectral clustering in the analysis of multi-layer categorical data. Furthermore, we also provide a highly efficient methodology for finding the optimal number of subgroups in multi-layer categorical data. To our knowledge, our work represents an early attempt to study the latent class analysis problem for multi-layer categorical data. Overall, this paper not only enhances the theoretical framework of latent class analysis in the context of multi-layer categorical data but also equips researchers and practitioners with practical and efficient tools for analyzing such complex datasets.

There are three major directions worth exploring in the future. First, designing rigorous methods to estimate $K$ in multi-layer categorical data is a promising and challenging problem. Second, extending the current work to accommodate the case where latent classes may overlap in multi-layer categorical data is appealing. Third, for multi-layer categorical data, exploring its dynamic nature, where different layers may have time-varying latent class structures, is interesting.
%\section*{CRediT authorship contribution statement}
%\textbf{Huan Qing:} Conceptualization; Data curation; Formal analysis; Funding acquisition; Methodology; Project administration; Resources; Software; Validation; Visualization; Writing-original draft; Writing-review $\&$ editing.
%\section*{Declaration of competing interest}
%The author declares no competing interests.
%\section*{Data availability}
%No actual data is utilized in this paper.
%\section*{Acknowledgements}
%H.Q. was sponsored by the Scientific Research Foundation of Chongqing University of Technology (Grant No: 0102240003) and
%the Natural Science Foundation of Chongqing, China (Grant No: CSTB2023NSCQ-LZX0048).
\appendix
\section{Proofs of theoretical results}\label{SecProofs}
\subsection{Proof of Lemma \ref{SVDEigenDecomposition}}
\begin{proof}
Since $\mathscr{R}_{l} = Z\Theta'_{l}$ for $l \in [L]$ and $Z'Z$ is a $K \times K$ non-singular matrix, we have $\Theta_{l}Z' = \mathscr{R}'_{l}$, which implies $\Theta_{l} = \mathscr{R}'_{l}Z(Z'Z)^{-1}$.

When $\mathrm{rank}(\sum_{l \in [L]}\Theta_{l}) = K$, we have $\mathscr{R}_{\mathrm{sum}} = \sum_{l \in [L]}\mathscr{R}_{l} = Z(\sum_{l \in [L]}\Theta_{l})' = U\Lambda B'$, which gives $U = Z(\sum_{l \in [L]}\Theta_{l})'B\Lambda^{-1} = ZX$, where $X = (\sum_{l \in [L]}\Theta_{l})'B\Lambda^{-1}$. Since $\mathrm{rank}(\sum_{l \in [L]}\Theta_{l}) = K$, and $\mathrm{rank}(B) = K$, we have $\mathrm{rank}(X) = K$. Combining $U'U = I_{K \times K}$ and $U = ZX$ gives $(ZX)'ZX = I_{K \times K} \Rightarrow X'Z'ZX = I_{K \times K} \Rightarrow XX' = (Z'Z)^{-1}$. Since $Z'Z$ is diagonal with $(Z'Z)(k,k)=N_{k}$ for $k\in[K]$, Equation (\ref{XDistance}) holds by $XX' = (Z'Z)^{-1}$.

When $\mathrm{rank}(\sum_{l \in [L]}\Theta'_{l}\Theta_{l}) = K$, we have $\mathcal{S}_{\mathrm{sum}} = \sum_{l \in [L]}\mathscr{R}_{l}\mathscr{R}'_{l} = Z(\sum_{l \in [L]}\Theta'_{l}\Theta_{l})Z' = V\Sigma V'$, which gives $V = Z(\sum_{l \in [L]}\Theta'_{l}\Theta_{l})Z'V\Sigma^{-1} = ZY$, where $Y = (\sum_{l \in [L]}\Theta'_{l}\Theta_{l})Z'V\Sigma^{-1}$. Since $\mathrm{rank}(\sum_{l \in [L]}\Theta'_{l}\Theta_{l}) = K$, $\mathrm{rank}(Z) = K$, and $\mathrm{rank}(V) = K$, we have $\mathrm{rank}(Y) = K$. Equation (\ref{YDistance}) holds similarly to Equation (\ref{XDistance}).
\end{proof}
\subsection{Proof of Lemma \ref{boundSumDeSumMLLCM}}
\begin{proof}
Let $e_{i}$ be a vector such that $e_{i}(i)=1$ and $e_{i}(\bar{i})=0$ for $i\in[N], \bar{i}\neq i, \bar{i}\in[N]$. Let $\tilde{e}_{j}$ be a $J\times 1$ vector such that $\tilde{e}_{j}(j)=1$ and $\tilde{e}_{j}(\bar{j})=0$ for $j\in[J], \bar{j}\neq j, \bar{j}\in[J]$. Let $W=R_{\mathrm{sum}}-\mathscr{R}_{\mathrm{sum}}$. We have $W=\sum_{l\in[L]}\sum_{i\in[N]}\sum_{j\in[J]}(R_{l}(i,j)-\mathscr{R}_{l}(i,j))e_{i}\tilde{e}'_{j}$. Let $W^{(ijl)}=(R_{l}(i,j)-\mathscr{R}_{l}(i,j))e_{i}\tilde{e}'_{j}$ for $i\in[N], j\in[J],l\in[L]$. We have $W=\sum_{l\in[L]}\sum_{i\in[N]}\sum_{j\in[J]}W^{(ijl)}$. For $W^{(ijl)}$, we have
\begin{itemize}
  \item $\mathbb{E}(W^{(ijl)})=0$ since $\mathbb{E}(R_{l})=\mathscr{R}_{l}$ for $l\in[L], i\in[N], j\in[J]$.
  \item $\|W^{(ijl)}\|=\|(R_{l}(i,j)-\mathscr{R}_{l}(i,j))e_{i}\tilde{e}'_{j}\|\leq M$ for $ l\in[L],i\in[N], j\in[J]$.
  \item $\mathbb{E}((R_{l}(i,j)-\mathscr{R}_{l}(i,j))^{2})=\mathrm{Var}(R_{l}(i,j))=M\frac{\mathscr{R}_{l}(i,j)}{M}(1-\frac{\mathscr{R}_{l}(i,j)}{M})=\mathscr{R}_{l}(i,j)(1-\frac{\mathscr{R}_{l}(i,j)}{M})\leq\mathscr{R}_{l}(i,j)\leq\rho$ since $R_{l}(i,j)\sim\mathrm{Binomial}(M,\frac{\mathscr{R}_{l}(i,j)}{M})$ for $l\in[L], i\in[N], j\in[J]$.
  \item Set $\sigma^{2}=\mathrm{max}(\|\sum_{l\in[L]}\sum_{i\in[N]}\sum_{j\in[J]}\mathbb{E}(W^{(ijl)}(W^{(ijl)})')\|,\|\sum_{l\in[L]}\sum_{i\in[N]}\sum_{j\in[J]}\mathbb{E}((W^{(ijl)})'W^{(ijl)})\|)$. We have
      \begin{align*}
      \|\sum_{l\in[L]}\sum_{i\in[N]}\sum_{j\in[J]}\mathbb{E}(W^{(ijl)}(W^{(ijl)})')\|&=\|\sum_{l\in[L]}\sum_{i\in[N]}\sum_{j\in[J]}\mathbb{E}((R_{l}(i,j)-\mathscr{R}_{l}(i,j))^{2}e_{i}e'_{i})\|=\mathrm{max}_{i\in[N]}\sum_{l\in[L]}\sum_{j\in[J]}\mathbb{E}((R_{l}(i,j)-\mathscr{R}_{l}(i,j))^{2})\leq\rho JL.
      \end{align*}
      Similarly, we have $
      \|\sum_{l\in[L]}\sum_{i\in[N]}\sum_{j\in[J]}\mathbb{E}((W^{(ijl)})'W^{(ijl)})\|\leq\rho NL.$ Therefore, $\sigma^{2}\leq \rho L\mathrm{max}(J,N)$.
\end{itemize}

Set $t=\frac{\sqrt{\alpha+1}(\sqrt{\alpha+1}+\sqrt{\alpha+19})}{3}\sqrt{\rho L\mathrm{max}(J,N)\mathrm{log}(J+N+L)}$ for any $\alpha\geq0$. Combining Assumption \ref{Assum1} with Theorem 1.6 in \citep{tropp2012user} gives
\begin{align*}
\mathbb{P}(\|W\|\geq t)\leq(J+N)\cdot\mathrm{exp}(\frac{-t^{2}/2}{\sigma^{2}+\frac{Mt}{3}})&\leq(J+N)\cdot\mathrm{exp}(\frac{-t^{2}/2}{\rho L\mathrm{max}(J,N)+\frac{Mt}{3}})\\
&=(J+N)\cdot\mathrm{exp}(-(\alpha+1)\mathrm{log}(J+N+L)\frac{1}{\frac{18}{(\sqrt{\alpha+1}+\sqrt{\alpha+19})^{2}}+\frac{2\sqrt{\alpha+1}}{\sqrt{\alpha+1}+\sqrt{\alpha+19}}\sqrt{\frac{M^{2}\mathrm{log}(J+N+L)}{\rho L\mathrm{max}(J,N)}}})\\
&\leq (J+N)\cdot\mathrm{exp}(-(\alpha+1)\mathrm{log}(J+N+L)\frac{1}{\frac{18}{(\sqrt{\alpha+1}+\sqrt{\alpha+19})^{2}}+\frac{2\sqrt{\alpha+1}}{\sqrt{\alpha+1}+\sqrt{\alpha+19}}})\\
&=(J+N)\cdot\mathrm{exp}(-(\alpha+1)\mathrm{log}(J+N+L))=\frac{J+N}{(J+N+L)^{\alpha+1}}\leq\frac{1}{(J+N+L)^{\alpha}}
\end{align*}
Hence,
\begin{align*}
\|R_{\mathrm{sum}}-\mathscr{R}_{\mathrm{sum}}\|=O(\sqrt{\rho L\mathrm{max}(N,J)\mathrm{log}(N+J+L)}).
\end{align*}

Define an $N\times N$ matrix $E^{(i\bar{i})}$ such that its $(i,\bar{i})$-th entry equals 1 while all the other elements equal 0 for $i\in[N],\bar{i}\in[N]$. For $\|\tilde{S}_{\mathrm{sum}}-\mathcal{S}_{\mathrm{sum}}\|$, we have
\begin{align*}
\|\tilde{S}_{\mathrm{sum}}-\mathcal{S}_{\mathrm{sum}}\|&=\|\sum_{l\in[L]}(R_{l}R'_{l}-D_{l}-\mathscr{R}_{l}\mathscr{R}'_{l})=\sum_{l\in[L]}\sum_{i\in[N]}\sum_{\bar{i}\in[N]}\sum_{j\in[N]}(R_{l}(i,j)R_{l}(\bar{i},j)-\mathscr{R}_{l}(i,j)\mathscr{R}_{l}(\bar{i},j))E^{(i\bar{i})}-\sum_{l\in[L]}D_{l}\|\\
&=\|\sum_{l\in[N]}\sum_{i\in[N]}\sum_{\bar{i}\neq i, \bar{i}\in[N]}\sum_{j\in[J]}(R_{l}(i,j)R_{l}(\bar{i},j)-\mathscr{R}_{l}(i,j)\mathscr{R}_{l}(\bar{i},j))E^{(i\bar{i})}+\sum_{l\in[L]}\sum_{i\in[N]}\sum_{j\in[J]}(R^{2}_{l}(i,j)-\mathscr{R}^{2}_{l}(i,j))E^{(ii)}-\sum_{l\in[L]}D_{l}\|\\
&=\|\sum_{l\in[N]}\sum_{i\in[N]}\sum_{\bar{i}\neq i, \bar{i}\in[N]}\sum_{j\in[J]}(R_{l}(i,j)R_{l}(\bar{i},j)-\mathscr{R}_{l}(i,j)\mathscr{R}_{l}(\bar{i},j))E^{(i\bar{i})}-\sum_{l\in[L]}\sum_{i\in[N]}\sum_{j\in[J]}\mathscr{R}^{2}_{l}(i,j)E^{(ii)}\|\\
&\leq\|\sum_{l\in[N]}\sum_{i\in[N]}\sum_{\bar{i}\neq i, \bar{i}\in[N]}\sum_{j\in[J]}(R_{l}(i,j)R_{l}(\bar{i},j)-\mathscr{R}_{l}(i,j)\mathscr{R}_{l}(\bar{i},j))E^{(i\bar{i})}\|+\|\sum_{l\in[L]}\sum_{i\in[N]}\sum_{j\in[J]}\mathscr{R}^{2}_{l}(i,j)E^{(ii)}\|\\
&=\|\sum_{l\in[N]}\sum_{i\in[N]}\sum_{\bar{i}\neq i, \bar{i}\in[N]}\sum_{j\in[J]}(R_{l}(i,j)R_{l}(\bar{i},j)-\mathscr{R}_{l}(i,j)\mathscr{R}_{l}(\bar{i},j))E^{(i\bar{i})}\|+\mathrm{max}_{i\in[N]}\sum_{l\in[L]}\sum_{j\in[J]}\mathscr{R}^{2}_{l}(i,j)\\
&\leq\|\sum_{l\in[N]}\sum_{i\in[N]}\sum_{\bar{i}\neq i, \bar{i}\in[N]}\sum_{j\in[J]}(R_{l}(i,j)R_{l}(\bar{i},j)-\mathscr{R}_{l}(i,j)\mathscr{R}_{l}(\bar{i},j))E^{(i\bar{i})}\|+\rho^{2}JL\\
\end{align*}
Set $\tilde{W}^{(i\bar{i}jl)}=(R_{l}(i,j)R_{l}(\bar{i},j)-\mathscr{R}_{l}(i,j)\mathscr{R}_{l}(\bar{i},j))E^{(i\bar{i})}$. For $l\in[N], i\in[N], \bar{i}\neq i, \bar{i}\in[N], j\in[N]$, we have :
\begin{itemize}
  \item $\mathbb{E}(\tilde{W}^{(i\bar{i}jl)})=0$ because $i\neq\bar{i}$.
  \item $\|\tilde{W}^{(i\bar{i}jl)}\|\leq M^{2}$.
\item Set $\tilde{\sigma}^{2}=\mathrm{max}(\|\sum_{l\in[L]}\sum_{i\in[N]}\sum_{\bar{i}\neq i, \bar{i}\in[N]}\sum_{j\in[J]}\mathbb{E}(\tilde{W}^{(i\bar{i}jl)}(\tilde{W}^{(i\bar{i}jl)})')\|,\|\sum_{l\in[L]}\sum_{i\in[N]}\sum_{\bar{i}\neq i, \bar{i}\in[N]}\sum_{j\in[J]}\mathbb{E}((\tilde{W}^{(i\bar{i}jl)})'\tilde{W}^{(i\bar{i}jl)})\|)$. We have
    \begin{align*}
    &\|\sum_{l\in[L]}\sum_{i\in[N]}\sum_{\bar{i}\neq i, \bar{i}\in[N]}\sum_{j\in[J]}\mathbb{E}(\tilde{W}^{(i\bar{i}jl)}(\tilde{W}^{(i\bar{i}jl)})')\|=\|\sum_{l\in[L]}\sum_{i\in[N]}\sum_{\bar{i}\neq i, \bar{i}\in[N]}\sum_{j\in[J]}\mathbb{E}((R_{l}(i,j)R_{l}(\bar{i},j)-\mathscr{R}_{l}(i,j)\mathscr{R}_{l}(\bar{i},j))^{2}E^{(ii)})\|\\
    &=\|\sum_{l\in[L], i\in[N], \bar{i}\neq i,\bar{i}\in[N], j\in[J]}E^{(ii)}(\mathbb{E}(R^{2}_{l}(i,j)R^{2}_{l}(\bar{i},j))-\mathscr{R}^{2}_{l}(i,j)\mathscr{R}^{2}_{l}(\bar{i},j))\|\\
    &=\|\sum_{l\in[L], i\in[N], \bar{i}\neq i,\bar{i}\in[N], j\in[J]}E^{(ii)}((\mathscr{R}^{2}_{l}(i,j)+\mathrm{Var}(R_{l}(i,j)))(\mathscr{R}^{2}_{l}(\bar{i},j)+\mathrm{Var}(R_{l}(\bar{i},j)))-\mathscr{R}^{2}_{l}(i,j)\mathscr{R}^{2}_{l}(\bar{i},j))\|\\
    &=\mathrm{max}_{i\in[N]}\sum_{l\in[L],\bar{i}\neq i,\bar{i}\in[N], j\in[J]}((\mathscr{R}^{2}_{l}(i,j)+\mathrm{Var}(R_{l}(i,j)))(\mathscr{R}^{2}_{l}(\bar{i},j)+\mathrm{Var}(R_{l}(\bar{i},j)))-\mathscr{R}^{2}_{l}(i,j)\mathscr{R}^{2}_{l}(\bar{i},j))\\
    &=\mathrm{max}_{i\in[N]}\sum_{l\in[L],\bar{i}\neq i,\bar{i}\in[N], j\in[J]}(\mathscr{R}^{2}_{l}(i,j)\mathrm{Var}(R_{l}(\bar{i},j))+\mathscr{R}^{2}_{l}(\bar{i},j)\mathrm{Var}(R_{l}(i,j))+\mathrm{Var}(R_{l}(i,j))\mathrm{Var}(R_{l}(\bar{i},j)))\\
    &\leq \mathrm{max}_{i\in[N]}\sum_{l\in[L],\bar{i}\neq i,\bar{i}\in[N], j\in[J]}(2\rho^{3}+\rho^{2})\leq (2\rho^{3}+\rho^{2})JNL.
    \end{align*}
\end{itemize}
Similarly, $\|\sum_{l\in[L]}\sum_{i\in[N]}\sum_{\bar{i}\neq i, \bar{i}\in[N]}\sum_{j\in[J]}\mathbb{E}((\tilde{W}^{(i\bar{i}jl)})'\tilde{W}^{(i\bar{i}jl)})\|\leq(2\rho^{3}+\rho^{2})JNL$. Hence, we have $\tilde{\sigma}^{2}\leq (2\rho^{3}+\rho^{2})JNL$. Set $\tilde{t}=\frac{\sqrt{\alpha+1}(\sqrt{\alpha+1}+\sqrt{\alpha+19})}{3}\sqrt{(2\rho^{3}+\rho^{2}) NJL\mathrm{log}(J+N+L)}$ for any $\alpha\geq0$. By Theorem 1.6 of \citep{tropp2012user}, Assumption \ref{Assum2}, and  $O(2\rho^{3}+\rho^{2})=O(\rho^{2}(1+2\rho))=O(\rho^{2})$ since we allow $\rho$ to decrease to 0, we get
\begin{align*}
\mathbb{P}(\|\sum_{l\in[L]}\sum_{i\in[N]}\sum_{\bar{i}\neq i,\bar{i}\in[N]}\sum_{j\in[J]}\tilde{W}^{(i\bar{i}jl)}\|\geq\tilde{t})&\leq 2N\cdot\mathrm{exp}(\frac{-\tilde{t}^{2}/2}{\tilde{\sigma}^{2}+\frac{M^{2}\tilde{t}}{3}})\leq 2N\cdot\mathrm{exp}(\frac{-\tilde{t}^{2}/2}{(2\rho^{3}+\rho^{2})JNL+\frac{M^{2}\tilde{t}}{3}})\\
&=2N\cdot\mathrm{exp}(-(\alpha+1)\mathrm{log}(J+N+L)\frac{1}{\frac{18}{(\sqrt{\alpha+1}+\sqrt{\alpha+19})^{2}}+\frac{2\sqrt{\alpha+1}}{\sqrt{\alpha+1}+\sqrt{\alpha+19}}\sqrt{\frac{M^{4}\mathrm{log}(J+N+L)}{(2\rho^{3}+\rho^{2})JNL}}})\\
&\leq\frac{2N}{(J+N+L)^{\alpha+1}}\leq\frac{2}{(J+N+L)^{\alpha}}.
\end{align*}
Hence,
\begin{align*}
\|\tilde{S}_{\mathrm{sum}}-\mathcal{S}_{\mathrm{sum}}\|=O(\sqrt{\rho^{2}NJL\mathrm{log}(J+N+L)})+\rho^{2}JL.
\end{align*}
Since $S_{\mathrm{sum}}-\mathcal{S}_{\mathrm{sum}}=\tilde{S}_{\mathrm{sum}}-\mathcal{S}_{\mathrm{sum}}+\sum_{l\in[L]}D_{l}$, we have $\|S_{\mathrm{sum}}-\mathcal{S}_{\mathrm{sum}}\|\leq\|\tilde{S}_{\mathrm{sum}}-\mathcal{S}_{\mathrm{sum}}\|+\|\sum_{l\in[L]}D_{l}\|\leq\|\tilde{S}_{\mathrm{sum}}-\mathcal{S}_{\mathrm{sum}}\|+M^{2}JL$. Setting $\alpha=3$, this lemma holds.
\end{proof}
\subsection{Proof of Theorem \ref{mainMLLCM}}
\begin{proof}
Since $\mathscr{R}_{\mathrm{sum}}=\rho Z(\sum_{l\in[L]}B_{l})'$, we have
\begin{align}\label{Reig}
\sigma_{K}(\mathscr{R}_{\mathrm{sum}})=|\lambda_{K}(\mathscr{R}_{\mathrm{sum}})|=\rho|\lambda_{K}(Z(\sum_{l\in[L]}B_{l})')|\geq\rho\lambda_{K}(Z)|\lambda_{K}(\sum_{l\in[L]}B_{l})|=\rho \sqrt{N_{\mathrm{min}}}\sigma_{K}(\sum_{l\in[L]}B_{l}).
\end{align}
Since $\mathcal{S}_{\mathrm{sum}}=\rho^{2}Z(\sum_{l\in[L]}B'_{l}B_{l})Z'$, we have
\begin{align}\label{Seig}
\sigma_{K}(\mathcal{S}_{\mathrm{sum}})=\rho^{2}|\lambda_{K}(Z(\sum_{l\in[L]}B'_{l}B_{l})Z')|\geq\rho^{2}\lambda_{K}(Z'Z)|\lambda_{K}(\sum_{l\in[L]}B'_{l}B_{l})|=\rho^{2}N_{\mathrm{min}}\sigma_{K}(\sum_{l\in[L]}B'_{l}B_{l}).
\end{align}
Through the analysis presented in the proof of Lemma 3 from \citep{zhou2019analysis}, there exist two orthogonal matrices, denoted as $\mathcal{O}$ and $\tilde{\mathcal{O}}$, which satisfy
\begin{align*}
\|\hat{U}\mathcal{O}-U\|_{F}\leq\frac{2\sqrt{2K}\|R_{\mathrm{sum}}-\mathscr{R}_{\mathrm{sum}}\|}{\sigma_{K}(\mathscr{R}_{\mathrm{sum}})}\mathrm{~and~}\|\hat{V}\tilde{\mathcal{O}}-V\|_{F}\leq\frac{2\sqrt{2K}\|\tilde{S}_{\mathrm{sum}}-\mathcal{S}_{\mathrm{sum}}\|}{\sigma_{K}(\mathcal{S}_{\mathrm{sum}})}.
\end{align*}
By Equations (\ref{Reig}) and (\ref{Seig}), we have
\begin{align}\label{boundUV}
\|\hat{U}\mathcal{O}-U\|_{F}\leq\frac{2\sqrt{2K}\|R_{\mathrm{sum}}-\mathscr{R}_{\mathrm{sum}}\|}{\rho\sqrt{N_{\mathrm{min}}}\sigma_{K}(\sum_{l\in[L]}B_{l})}\mathrm{~and~}\|\hat{V}\tilde{\mathcal{O}}-V\|_{F}\leq\frac{2\sqrt{2K}\|\tilde{S}_{\mathrm{sum}}-\mathcal{S}_{\mathrm{sum}}\|}{\rho^{2}N_{\mathrm{min}}\sigma_{K}(\sum_{l\in[L]}B'_{l}B_{l})}.
\end{align}
Drawing upon Lemma 2 from \citep{joseph2016impact}, Equation (\ref{XDistance}), and Equation (\ref{YDistance}), and employing a proof methodology similar to that utilized in Theorem 1 of \citep{qing2023community}, we obtain at the following result:
\begin{align*}
\hat{f}_{LCA-SoR}=O(\frac{KN_{\mathrm{max}}\|\hat{U}\mathcal{O}-U\|^{2}_{F}}{N_{\mathrm{min}}})\mathrm{~and~}\hat{f}_{LCA-DSoG}=O(\frac{KN_{\mathrm{max}}\|\hat{V}\tilde{\mathcal{O}}-V\|^{2}_{F}}{N_{\mathrm{min}}}).
\end{align*}
By Equation (\ref{boundUV}), we have
\begin{align*}
\hat{f}_{LCA-SoR}=O(\frac{K^{2}N_{\mathrm{max}}\|R_{\mathrm{sum}}-\mathscr{R}_{\mathrm{sum}}\|^{2}}{\rho^{2}N^{2}_{\mathrm{min}}\sigma^{2}_{K}(\sum_{l\in[L]}B_{l})})\mathrm{~and~}\hat{f}_{LCA-DSoG}=O(\frac{K^{2}N_{\mathrm{max}}\|\tilde{S}_{\mathrm{sum}}-\mathcal{S}_{\mathrm{sum}}\|^{2}}{\rho^{4}N^{3}_{\mathrm{min}}\sigma^{2}_{K}(\sum_{l\in[L]}B'_{l}B_{l})}).
\end{align*}
Similarly, we have
\begin{align*}
\hat{f}_{LCA-SoG}=O(\frac{K^{2}N_{\mathrm{max}}\|S_{\mathrm{sum}}-\mathcal{S}_{\mathrm{sum}}\|^{2}}{\rho^{4}N^{3}_{\mathrm{min}}\sigma^{2}_{K}(\sum_{l\in[L]}B'_{l}B_{l})}).
\end{align*}
By Lemma \ref{boundSumDeSumMLLCM}, Assumption \ref{Assum11}, and Assumption \ref{Assum22}, we have
\begin{align*}
&\hat{f}_{LCA-SoR}=O(\frac{K^{2}N_{\mathrm{max}}\mathrm{max}(J,N)\mathrm{log}(J+N+L)}{\rho N^{2}_{\mathrm{min}}JL}),\\
&\hat{f}_{LCA-DSoG}=O(\frac{K^{2}N_{\mathrm{max}}N\mathrm{log}(J+N+L)}{\rho^{2}N^{3}_{\mathrm{min}}JL})+O(\frac{K^{2}N_{\mathrm{max}}}{N^{3}_{\mathrm{min}}}),\\
&\hat{f}_{LCA-SoG}=O(\frac{K^{2}N_{\mathrm{max}}N\mathrm{log}(J+N+L)}{\rho^{2}N^{3}_{\mathrm{min}}JL})+O(\frac{K^{2}N_{\mathrm{max}}M^{4}}{N^{3}_{\mathrm{min}}\rho^{4}}).
\end{align*}
Set $Z(Z'Z)^{-1}=Q$ and $\hat{Z}(\hat{Z}'\hat{Z})^{-1}=\hat{Q}$. By Lemma \ref{boundSumDeSumMLLCM}, we have
\begin{align*}
\|\sum_{l\in[L]}(\hat{\Theta}_{l}-\Theta_{l})\|&=\|(\sum_{l\in[L]}R_{l})'\hat{Q}-(\sum_{l\in[L]}\mathscr{R}_{l})'Q\|=\|(R'_{\mathrm{sum}}-\mathscr{R}'_{\mathrm{sum}})\hat{Q}+\mathscr{R}'_{\mathrm{sum}}(\hat{Q}-Q)\|\\
&\leq\|(R_{\mathrm{sum}}-\mathscr{R}_{\mathrm{sum}})'\hat{Q}\|+\|\mathscr{R}'_{\mathrm{sum}}(\hat{Q}-Q)\|\\
&\leq\|R_{\mathrm{sum}}-\mathscr{R}_{\mathrm{sum}}\|\|\hat{Q}\|+\|\rho Z(\sum_{l\in[L]}B_{l})'\|\|\hat{Q}-Q\|\\
&\leq\|R_{\mathrm{sum}}-\mathscr{R}_{\mathrm{sum}}\|\|\hat{Q}\|+\rho \|Z\|\|\sum_{l\in[L]}B_{l}\|\|\hat{Q}-Q\|\\
&=O(\sqrt{\frac{\rho L\mathrm{max}(N,J)\mathrm{log}(N+J+L)}{N_{\mathrm{min}}}})+O(\frac{\rho\sqrt{N_{\mathrm{max}}}\sigma_{1}(\sum_{l\in[L]}B_{l})}{\sqrt{N_{\mathrm{min}}}}),
\end{align*}
where $\frac{1}{\sqrt{N_{\mathrm{min}}}}$ is used as estimations of $\|\hat{Q}\|$ and $\|\hat{Q}-Q\|$ due to the fact that $\|Z(Z'Z)^{-1}\|=\frac{1}{\sqrt{N_{\mathrm{min}}}}$ and $\hat{Z}$ is an estimation of $Z$. Because $\|\sum_{l\in[L]}(\hat{\Theta}_{l}-\Theta_{l})\|_{F}\leq\sqrt{K}\|\sum_{l\in[L]}(\hat{\Theta}_{l}-\Theta_{l})\|$, we get
\begin{align*}
\|\sum_{l\in[L]}(\hat{\Theta}_{l}-\Theta_{l})\|_{F}=O(\sqrt{\frac{\rho KL\mathrm{max}(N,J)\mathrm{log}(J+N+L)}{N_{\mathrm{min}}}})+O(\frac{\rho\sqrt{KN_{\mathrm{max}}}\sigma_{1}(\sum_{l\in[L]}B_{l})}{\sqrt{N_{\mathrm{min}}}}),
\end{align*}
Combining Assumption \ref{Assum11} with the fact that $\|\sum_{l\in[L]}\Theta_{l}\|_{F}\geq\|\sum_{l\in[L]}\Theta_{l}\|=\rho\sigma_{1}(\sum_{l\in[L]}B_{l})$ gives
\begin{align*}
\frac{\|\sum_{l\in[L]}(\hat{\Theta}_{l}-\Theta_{l})\|_{F}}{\|\sum_{l\in[L]}\Theta_{l}\|_{F}}&\leq\frac{\|\sum_{l\in[L]}(\hat{\Theta}_{l}-\Theta_{l})\|_{F}}{\rho\sigma_{1}(\sum_{l\in[L]}B_{l})}=O(\sqrt{\frac{KL\mathrm{max}(N,J)\mathrm{log}(N+J+L)}{\rho N_{\mathrm{min}}\sigma^{2}_{1}(\sum_{l\in[L]}B_{l})}})+O(\sqrt{\frac{KN_{\mathrm{max}}}{N_{\mathrm{min}}}})\\
&\leq O(\sqrt{\frac{KL\mathrm{max}(N,J)\mathrm{log}(N+J+L)}{\rho N_{\mathrm{min}}\sigma^{2}_{K}(\sum_{l\in[L]}B_{l})}})+O(\sqrt{\frac{KN_{\mathrm{max}}}{N_{\mathrm{min}}}})\\
&=O(\sqrt{\frac{KL\mathrm{max}(N,J)\mathrm{log}(N+J+L)}{\rho N_{\mathrm{min}}\sigma^{2}_{K}(\sum_{l\in[L]}B_{l})}})=O(\sqrt{\frac{K\mathrm{max}(N,J)\mathrm{log}(N+J+L)}{\rho N_{\mathrm{min}}JL}}),
\end{align*}
where we do not consider the term $O(\sqrt{\frac{KN_{\mathrm{max}}}{N_{\mathrm{min}}}})$ since it does not absorb the two key model parameters $\rho$ and $L$.
\end{proof}
\subsection{Proof of Lemma \ref{CompareSoRDSoG}}
\begin{proof}
By Theorem \ref{mainMLLCM}, we have:
\begin{itemize}
  \item \textbf{Case 1}:  LCA-DSoG's error rate is at the order $\frac{K^{2}N_{\mathrm{max}}N\mathrm{log}(J+N+L)}{\rho^{2}N^{3}_{\mathrm{min}}JL}$. This cases requires
      \begin{align}\label{rho1}
       \frac{K^{2}N_{\mathrm{max}}N\mathrm{log}(J+N+L)}{\rho^{2}N^{3}_{\mathrm{min}}JL}\gg\frac{K^{2}N_{\mathrm{max}}}{N^{3}_{\mathrm{min}}}\Leftrightarrow \rho\ll\sqrt{\frac{N\mathrm{log}(J+N+L)}{JL}}.
      \end{align}
  For this case, we have $\hat{f}_{LCA-DSoG}=O(\frac{K^{2}N_{\mathrm{max}}N\mathrm{log}(J+N+L)}{\rho^{2}N^{3}_{\mathrm{min}}JL})$. When $\hat{f}_{LCA-DSoG}\ll\hat{f}_{LCA-SoR}$, we have
  \begin{align}\label{rho2}
  \frac{K^{2}N_{\mathrm{max}}\mathrm{max}(J,N)\mathrm{log}(J+N+L)}{\rho N^{2}_{\mathrm{min}}JL}\gg\frac{K^{2}N_{\mathrm{max}}N\mathrm{log}(J+N+L)}{\rho^{2}N^{3}_{\mathrm{min}}JL}\Leftrightarrow \rho\gg\frac{N}{N_{\mathrm{min}}\mathrm{max}(J,N)}.
  \end{align}
  Combing Equation (\ref{rho1}) with Equation (\ref{rho2}) gives
  \begin{align*}
  \sqrt{\frac{N\mathrm{log}(J+N+L)}{JL}}\gg\frac{N}{N_{\mathrm{min}}\mathrm{max}(J,N)}\Leftrightarrow L\ll\frac{N^{2}_{\mathrm{min}}\mathrm{max}(J^{2},N^{2})\mathrm{log}(J+N+L)}{JN}.
  \end{align*}
  \item \textbf{Case 2}:  LCA-DSoG's error rate is at the order $\frac{K^{2}N_{\mathrm{max}}}{N^{3}_{\mathrm{min}}}$. This cases requires
      \begin{align}\label{rho11}
       \frac{K^{2}N_{\mathrm{max}}N\mathrm{log}(J+N+L)}{\rho^{2}N^{3}_{\mathrm{min}}JL}\ll\frac{K^{2}N_{\mathrm{max}}}{N^{3}_{\mathrm{min}}}\Leftrightarrow \rho\gg\sqrt{\frac{N\mathrm{log}(J+N+L)}{JL}}.
      \end{align}
  For this case, when $\hat{f}_{LCA-DSoG}\ll\hat{f}_{LCA-SoR}$, we have
  \begin{align}\label{rho22}
  \frac{K^{2}N_{\mathrm{max}}\mathrm{max}(J,N)\mathrm{log}(J+N+L)}{\rho N^{2}_{\mathrm{min}}JL}\gg\frac{K^{2}N_{\mathrm{max}}}{N^{3}_{\mathrm{min}}}\Leftrightarrow\rho\ll\frac{N_{\mathrm{min}}\mathrm{max}(J,N)\mathrm{log}(J+N+L)}{JL}. \end{align}
  Combing Equation (\ref{rho11}) with Equation (\ref{rho22}) gives
  \begin{align*}
 \sqrt{\frac{N\mathrm{log}(J+N+L)}{JL}}\ll\frac{N_{\mathrm{min}}\mathrm{max}(J,N)\mathrm{log}(J+N+L)}{JL}\Leftrightarrow L\ll\frac{N^{2}_{\mathrm{min}}\mathrm{max}(J^{2},N^{2})\mathrm{log}(J+N+L)}{JN}.
  \end{align*}
\end{itemize}
\end{proof}
\bibliographystyle{model5-names}\biboptions{authoryear}%ESWA style
\bibliography{refMLLCM}
\end{document}